\documentclass{article} 
\usepackage[preprint]{colm2026_conference}

\usepackage{microtype}
\microtypesetup{expansion=false}
\usepackage{hyperref}
\usepackage{url}
\usepackage{xurl}
\usepackage{booktabs}
\usepackage[utf8]{inputenc}
\usepackage[T1]{fontenc}
\usepackage{amsmath,amssymb,amsfonts,amsthm}
\usepackage{mathtools}
\usepackage{multirow}
\usepackage{hyperref}
\usepackage{xcolor}
\usepackage{enumitem}
\usepackage{caption}
\usepackage{subcaption}
\usepackage{algorithm}
\usepackage{algpseudocode}
\usepackage{natbib}
\usepackage{xspace}
\usepackage{nicefrac}
\usepackage{bm}
\usepackage{microtype}
\usepackage{hyperref}
\usepackage{url}
\usepackage{booktabs}
\usepackage{enumitem}
\usepackage[utf8]{inputenc} 
\usepackage[T1]{fontenc}    
\usepackage{hyperref}       
\usepackage{url}            
\usepackage{booktabs}       
\usepackage{amsfonts}       
\usepackage{makecell}       
\usepackage{nicefrac}       
\usepackage{microtype}      
\usepackage{xcolor}         
\usepackage{amsmath}
\usepackage{bbm}
\usepackage{multirow}
\usepackage{graphicx}
\usepackage{graphicx}
\usepackage{float}
\usepackage{wrapfig}
\usepackage{siunitx}

\usepackage{subcaption}
\usepackage{arydshln}  
\usepackage{mdframed,lipsum,calc}
\usepackage{subcaption}
\usepackage{hyperref}
\usepackage{array}
\usepackage{algorithm}
\usepackage{algpseudocode}
\usepackage{rotating}
\newcommand{\W}{\mathbf{W}}
\newcommand{\U}{\mathbf{U}}
\newcommand{\V}{\mathbf{V}}
\newcommand{\Sig}{\boldsymbol{\Sigma}}
\newcommand{\dW}{\Delta\mathbf{W}}
\renewcommand{\P}{\mathbf{P}}
\newcommand{\R}{\mathbb{R}}
\newcommand{\norm}[1]{\left\|#1\right\|}
\newcommand{\fdiag}{f_{\mathrm{diag}}}

\newcommand{\rupd}{r_{\Delta}}

\newcommand{\col}{\mathrm{col}}

\newcommand{\tss}[1]{{\,\scriptsize(#1\%)}}

\DeclareMathOperator{\diag}{diag}

\usepackage{lineno}

\definecolor{darkblue}{rgb}{0, 0, 0.5}
\hypersetup{colorlinks=true, citecolor=darkblue, linkcolor=darkblue, urlcolor=darkblue}

\title{GRRR: The Geometry of Reshaping, Rotation, and Routing in Decoder LLM post-training}

\author{
Jianing Qi$^{1}$,~~Hao Tang$^{1}$,~~Zhigang Zhu$^{1,2}$\\ 
$^1$CUNY Graduate Center, $^2$The City College of New York\\
{\tt\small \{jqi@gc, htang@gc, zzhu@ccny\}.cuny.edu}\\
}

\begin{document}

\ifcolmsubmission 
\linenumbers
\fi

\maketitle

\begin{abstract}
We study how post-training changes the weights of Large Language Models (LLMs) relative to their pretrained weights. Across 12 post-training chains with supervised fine-tuning (SFT) and reinforcement learning (RL), we express each weight update in the pretrained matrix’s singular value decomposition (SVD) frame. This decomposition separates the changes of three geometrically distinct components: diagonal values, which reshapes singular values; off-diagonal values, which rotates the coupling between pretrained input and output directions; and null-space values, which routes outside the matrix’s original nonzero SVD core. On a math evaluation suite, we find that removing the diagonal component usually preserves most of the gains from post-training. These results suggest that post-training gains are carried primarily by reconfiguring and extending pretrained pathways rather than by substantially changing singular values of pre-trained models.

\end{abstract} 

\section{Introduction}
\label{sec:intro}

Large-scale pretraining establishes much of an LLM’s general capability before post-training begins, and increasing pre-training scale improves loss and downstream performance \citep{NEURIPS2020_1457c0d6, kaplan2020scalinglawsneurallanguage, wei2022emergent}. These findings suggest that broad knowledge and computational capabilities are already encoded in the pretrained weights. Post-training can significantly change model behavior, such as instruction following, reasoning, and human alignment \citep{zhou2023lima, ouyang2022training, wu-etal-2024-language}. 
Despite these large behavioral changes, prior work suggests that post-training often acts as a relatively small perturbation around the pretrained model weight space rather than a full rewrite \citep{radiyadixit2020finefinetuningbelearning, tanwar2025understandingeffectsdomainfinetuning, mukherjee2025reinforcementlearningfinetunessmall}. 
Recent works investigate this question through singular value decomposition (SVD) analyses of pretrained weights and their updates \citep{zhu2025pathtakenrlvrprovably, he2026understandingposttrainingstructuralchanges}.
In this paper, we ask a more direct question: where do the functionally useful changes introduced by post-training live geometrically in the pre-trained weights space?

Existing spectral analyses of SVD reveal that post-training does more than simply change singular values. \citet{zhu2025pathtakenrlvrprovably} study Reinforcement Learning with Verifiable Rewards (RLVR) training dynamics and find that its weight matrices are steered away from the principal directions of the pretrained weights. In addition, \citet{he2026understandingposttrainingstructuralchanges} compare final checkpoints with pre-trained models and find near uniform singular value scaling together with coordinated changes of singular vectors. 
It naturally raises a follow-up question: in the final weight difference, which parts are functionally necessary for the behavior of post-trained model? Understanding where useful weight changes occur can guide the design of more interpretable and efficient post-training methods that target the components responsible for performance gains.
To answer this question, we project the post-trained weight updates onto the pretrained model's SVD coordinates, each weight update projects into SVD's three components: pretrained singular values basis, which are diagonal spectral scaling (\textbf{reshaping}); mixture of $U$ and $V$ rotation basis, which are off-diagonal mixing within the pretrained subspace (\textbf{rotation}); and those outside of SVD basis, which are null space route outside SVD basis (\textbf{routing}). We call this framework the \textbf{G}eometry of \textbf{r}eshaping, \textbf{r}otation, and \textbf{r}outing (\textbf{GRRR}).

Our main result shows that post-training gains do not rely on changing pretrained singular values in reasoning setting. Even when aggressive post-training puts more mass in the diagonal channel, removing that channel usually does not hurt downstream performance. GRRR also makes explicit the importance of off-diagonal (rotation) and null space (routing) updates -- a substantial part of the update can lie outside the pretrained SVD frame. The useful signal sits mainly in rotation and routing.

We establish this result through causal reconstruction of the partial final weight update for a tested model. For each post-training transition, we selectively add the reshaping, rotation, and routing components to the pretrained checkpoint and evaluate the resulting model.
On the 25 transitions where the full model improves over the base model by at least 1 point on the five-benchmark average, keeping only rotation and routing stays within 4.3 points of the full post-trained model in 24 cases. In our setting, the final weight difference changes behavior mainly by remixing pretrained features and routing beyond the pretrained frame, not by scaling singular values.

\paragraph{Contributions.}
We make three contributions. First, we introduce GRRR, a decomposition of post-training weight differences into diagonal reshaping, rotation, and routing. GRRR extends SVD-based weight analysis from descriptive geometry to downstream performance: by adding each component back to the pretrained model separately, we can test which components actually carry the performance gain. Second, we use that causal view to show that removing the diagonal from the final weight difference usually does not hurt on our shared evaluation suite, with one clear outlier involving heavy distillation. In most of the settings we study, the useful signal sits mainly in rotation and routing. Third, we show that the geometry is structured: incremental RL rotations are nearly orthogonal to the preceding SFT rotations, while weight updates produced by independently trained RL and SFT from the same base show weak alignment. Other than that, we do not see clear algorithm label differences reflected on weights difference.


\paragraph{Study scope.}
Our empirical scope is specific. All models in the paper are decoder LLMs. The main causal evaluation uses a five benchmark average dominated by math reasoning. We therefore limit our strongest claims to decoder LLM post-training in these settings, rather than to all post-training regime.


\section{Related Work}
\label{sec:related}

\textbf{Spectral Analysis of Neural Network Weights.}
SVD has long been used to study neural network training. \citet{martin2018implicitselfregularizationdeepneural} use random matrix theory to characterize how weight spectra evolve across training. \citet{yunis2024approachingdeeplearningspectral} extend this view to modern architectures and track both singular values and singular vectors. For post-training, \citet{he2026understandingposttrainingstructuralchanges} analyze full weight matrices and argue that post-training often looks like near uniform singular value scaling, akin to a temperature shift, together with coordinated rotations of singular vectors. \citet{zhu2025pathtakenrlvrprovably} study RLVR training dynamics through leading singular components and argue that RL is steered away from principal or diagonal directions while SFT more often targets them. GRRR is complementary to both. We decompose the final update $\Delta W$ in the pretrained basis rather than comparing full weights in their own coordinates. That keeps one shared frame across methods, supports add/remove causal ablations on the final checkpoint, and separates a null term outside the pretrained frame that full matrix SVD comparisons leave implicit.

\textbf{SVD-Based Parameter Efficient Fine Tuning.}
A parallel line of work uses the pretrained SVD basis to design efficient adapters. LoRA~\citep{hu2022lora} constrains updates to low rank. PiSSA, MiLoRA, and DoRA all use SVD structure in different ways~\citep{meng2024pissa, wang-etal-2025-milora, liu2024dora}.
SVFT~\citep{lingam2024svft} is closest to our setup. It parameterizes the weight update in the pretrained SVD basis and finds that training the full singular value matrix can outperform other LoRA configurations. Our goal is different. We do not use the decomposition for efficiency. We use it to ask which geometric part of the update carries the behavioral change.

\textbf{post-training Weight Changes.}
Several works study the structure of post-training weight changes. \citet{ilharco2023editing} define task vectors and find low alignment across tasks. \citet{shuttleworth2025lora} show that LoRA can introduce new directions outside the pretrained spectrum. \citet{si2025weightspectrainducedefficient} find that fine-tuning changes vector orientation more than singular values. On the SFT versus RL question, \citet{mukherjee2025reinforcement} show that RL updates only 5--30\% of parameters across seven algorithms, while \citet{jin2025rlfinetuninghealsood} argue that singular vector direction matters more than singular values for out of distribution (OOD) performance. GRRR ties these observations together. The diagonal matters less than it appears to, and the split between the off-diagonal term and the null space term helps explain why RL can look sparse while SFT looks larger.

\section{The GRRR Framework}
\label{sec:framework}

Existing spectral analyses decompose each trained model in its own coordinates which makes relative differences very difficult to compare. To make post-trained models comparable, we use the pretrained model as the shared coordinate system. This makes different post-trained models starting from the same pretrained model comparable with a same pretrained model coordinate. This gives one basis for every update and splits the change into three parts: diagonal reshaping, off-diagonal remixing, and null space routing. GRRR is a shared geometric decomposition that makes these updates comparable and supports direct causal tests.

\paragraph{The SVD Decomposition of Weight Changes}
\label{sec:svd}

We can decompose the pre-trained weight matrix $\W \in \R^{m \times n}$ with a thin SVD:
\begin{equation}
  \W = \U \Sig \V^\top,
  \quad \U \in \R^{m \times r},\; \Sig = \diag(\sigma_1,\dots,\sigma_r),\;
  \V \in \R^{n \times r},
\end{equation}
where $r = \min(m,n)$ and $\sigma_1 \ge \sigma_2 \ge \cdots \ge \sigma_r \ge 0$.
We view each singular triple $(u_i, \sigma_i, v_i)$ as one route from an input pattern to an output pattern. The singular value $\sigma_i$ says how strong that route is. Changing the diagonal part in $\Sig$ changes route strength, but changing the off-diagonal of $\Sig$ to none-zero means it changes how inputs are matched to outputs of $\U$ or $\V$. We use the \emph{thin} SVD of the pretrained matrix because it gives a shared orthonormal frame with $r=\min(m,n)$ for every update built from the same base, and we can nicely separate out the basis within the pretrained coordinates and the null residual outside of SVD basis.

\paragraph{The Perturbation Matrix}
\label{sec:pmatrix}

Given a pretrained base weight $\W_0 = \U_0 \Sig_0 \V_0^\top$ and a post-trained weight $\W_1$, we can express the update as $\dW = \W_1 - \W_0$. We project the update into the pretrained SVD coordinate as \textbf{perturbation matrix} because pretrained model is shared across many different post-trained models:
\begin{equation}
  \P = \U_0^\top (\dW) \V_0 \;\in\; \R^{r \times r}.
  \label{eq:pmatrix}
\end{equation}

The matrix $\P$ is the update written in the base model's SVD coordinates. Equivalently,
\[
\U_0^\top \W_1 \V_0 = \Sig_0 + \P,
\]
and we can see very clearly $ \W_1  =\U_0 (\Sig_0 + \P)\V_0^\top$. So $\P$ tells us how the post-training weight departs from the base model inside the pretrained frame. Then we build three components:
\begin{align}
  \dW_{\mathrm{diag}} &= \U_0 \diag(\P) \V_0^\top, \\
  \dW_{\mathrm{off}}  &= \U_0 \bigl(\P - \diag(\P)\bigr) \V_0^\top, \\
  \dW_{\mathrm{null}} &= \dW - \U_0 \P \V_0^\top.
\end{align}

These correspond to three simple kinds of change:
(1)  $\dW_{\mathrm{diag}}$: \textbf{diagonal spectral reshaping}. Changes singular value
        magnitudes within the pretrained subspace.
(2)  $\dW_{\mathrm{off}}$: \textbf{off-diagonal rotation}. Rotates pretrained singular directions
        within the same subspace (off-diagonal structure in $\P$).
(3)  $\dW_{\mathrm{null}}$: \textbf{null space routing}. The residual
        outside the pretrained SVD frame. While $\U_0 \P \V_0^\top$ is the
        orthogonal projection of $\dW$ onto the pretrained subspace,
        $\dW_{\mathrm{null}}$ captures all components that cannot be expressed
        in that basis.

For readability, we refer to $\dW_{\mathrm{off}}$ as \emph{rotation} and $\dW_{\mathrm{null}}$ as \emph{routing} throughout.


By construction, these three components are exactly orthogonal:
\begin{equation}
  \norm{\dW}_F^2
  = \norm{\dW_{\mathrm{diag}}}_F^2
  + \norm{\dW_{\mathrm{off}}}_F^2
  + \norm{\dW_{\mathrm{null}}}_F^2.
\end{equation}

This three way split matters because the pretrained SVD gives a shared reference frame across post-training methods. In that frame, the diagonal term rescales existing routes, the off-diagonal term mixes them, and the null term leaves the in frame part of the layer. We can use the decomposition for causal tests by adding each part back separately and checking which one actually carries the gain. Under the thin SVD definition above, when a weight matrix is square, $\U_0$ and $\V_0$ span the ambient spaces and the null term vanishes. In many transformer weights, however, the projections are rectangular and null part is significant.


\paragraph{The Interpretation of Components}
Each projection matrix $\W_0 \in \R^{m \times n}$ maps hidden states to outputs via $\W_0 h = \U_0 \Sig_0 (\V_0^\top h)$.
From a different view, we can treat the right singular vectors as specifying which hidden state directions this layer reads from, the left singular vectors as specifying which output directions it writes into, and the singular values as determining the gain of these read-write routes. The diagonal and off-diagonal components both lie in
\[
\mathcal{S}=\{\U_0 A \V_0^\top : A\in\R^{r\times r}\},
\]
so they only rescale or remix routes within the pretrained singular vector frame. 
The $\dW_{\mathrm{diag}}$ changes rescale the strength of the singular values and the $\dW_{\mathrm{off}}$ changes the mixture of the read-write routes within the frame. Instead of mapping $v_i$ primarily to $u_i$, off-diagonal mixes $v_i$ into $u_j$ for $i \neq j $. So the layer reuses its pretrained features but mixes input features with different outputs. The $\dW_{\mathrm{null}}$ component captures updates outside the nonzero pretrained SVD core.

For square projections, $\dW_{\mathrm{null}}$ vanishes identically under the thin SVD definition because $\U_0$ and $\V_0$ span the full ambient spaces. For rectangular matrices, the null term is the part of the update that cannot be written as $\U_0 A \V_0^\top$, and therefore lies outside that pretrained frame. Concretely, $\dW_{\mathrm{null}}$ lets the layer read from hidden state directions outside $\col(\V_0)$ or write into output directions outside $\col(\U_0)$, depending on which side is rank deficient. As an interpretation, it means if the features are in the null space, the weight matrix has zero process of it and adds nothing on top of the residual. Changing in null simply means it allows the layer to read and write features that the pretrained layer ignored, and it allows the post-trained weight matrix to express pretrained matrix could not express.

We want to emphasize this as a local geometric statement about one layer rather than claiming the network as a whole creates new features. Residual paths may already represent or process the local null features in other layers. Our claim is that, relative to the pretrained map of this layer, $\dW_{\mathrm{null}}$ accesses directions outside the layer's original SVD basis. Empirically, post-training usually preserves stable rank and keeps the leading singular value structure highly aligned with the pretrained weight. This suggests that post-training can change where a layer reads and writes without greatly increasing the number of dominant modes. We analyze the effective rank in Appendix~\ref{app:rank-invariance}.





\subsection{Experimental Setup}
\label{sec:setup}

\textbf{Models and transitions.}
We study 12 training chains spanning 8 architecture families: Qwen~2.5, Qwen~3, Mistral, LLaMA~3, OLMo~2, OLMo~3, DeepSeek, and MiMo. Model sizes range from 1.5B to 14B parameters \citep{qwen2025qwen25technicalreport, yang2025qwen3technicalreport, jiang2023mistral7b, grattafiori2024llama3herdmodels, olmo20252olmo2furious, olmo2025olmo3, Guo_2025, shao2024deepseekmathpushinglimitsmathematical, xiaomi2025mimounlockingreasoningpotential}. The chain identifiers run from C1 to C12. The main inventory contains 29 incremental transitions. Some appendices also include cumulative base to checkpoint transitions, which brings the count to 42 transition instances. The post-training methods cover SFT, DPO, PPO, GRPO, GPPO, RAFT, PRIME, and OlmoRL \citep{rafailov2024directpreferenceoptimizationlanguage, schulman2017proximalpolicyoptimizationalgorithms, shao2024deepseekmathpushinglimitsmathematical, su2026klearreasoneradvancingreasoningcapability, zhang2025dpor1, cui2025processreinforcementimplicitrewards, olmo2025olmo3}. Domains range from math reasoning to general instruction following and agent tasks. The full inventory appears in Appendix~\ref{app:transitions}, Tables~\ref{tab:chains}--\ref{tab:transitions}.

\textbf{Decomposition scope.}
For each transition we decompose all seven projection matrices ($q$, $k$, $v$, $o$, gate, up, down) in every transformer layer. Each matrix uses its own thin SVD of the pretrained weight, so the diagonal, off-diagonal, and null components are defined matrix by matrix. In the causal experiments, each component is added back independently to its corresponding base matrix, and the resulting model applies that replacement across all decomposed matrices. Unlike prior work that focuses on top-$K$ structure, we use the full decomposition of each matrix.

\textbf{Causal evaluation.}
For the causal decomposition in Section~\ref{sec:causal_decomp}, we build four ablated variants per transition: diagonal only, off only, null only, and an OffNull variant that keeps the off and null terms together. We compare them against the full update on five math benchmarks: MATH500, AIME2024, AMC2023, Minerva-Math, and OlympiadBench, all at temperature~0 using \citet{zhangonline}. We use the same evaluation harness for every ablation and report the unweighted average across the five benchmarks. For general instruction checkpoints, this should be read as recovery on a shared math-heavy probe suite rather than on the checkpoint's original training objective. Appendix~\ref{app:causal-eval} reports 28 transition level decompositions (11 SFT, 17 RL) across 11 chains. We exclude C4 web research agent since it is unrelated to math benchmarks.

We report all transitions in the appendix tables and raw plots. When we summarize \emph{recovered improvement}, we use only transitions where the full post-trained model improves over base by at least 1 point on the five benchmark average. This avoids unstable percentages near zero gain and separates genuine recovery of improvement from flat or negative gain cases. Under this rule, the informative set has 25 transitions. C10 SFT, C5 zephyr-DPO, and C7 RLVR remain in the appendix for completeness but are not used as headline evidence about recovered gain.
 
\section{The Diagonal Usually Does Not Improve the Performance}
\label{sec:diagonal}

We start with the most common spectral question. Does post-training work by reshaping the pretrained spectrum, and does that reshaping matter? In most cases, no. Across models, the singular value profile before and after post-training is almost unchanged, even when behavior changes a lot. Removing the diagonal spectral scaling term also usually leaves performance nearly intact.

\subsection{post-training Usually Preserves the Spectrum}
\label{sec:spectrum_preserve}

\begin{figure}[ht]
\vspace{-0.2cm}
  \centering
  \begin{subfigure}[b]{0.48\textwidth}
    \includegraphics[width=\textwidth]{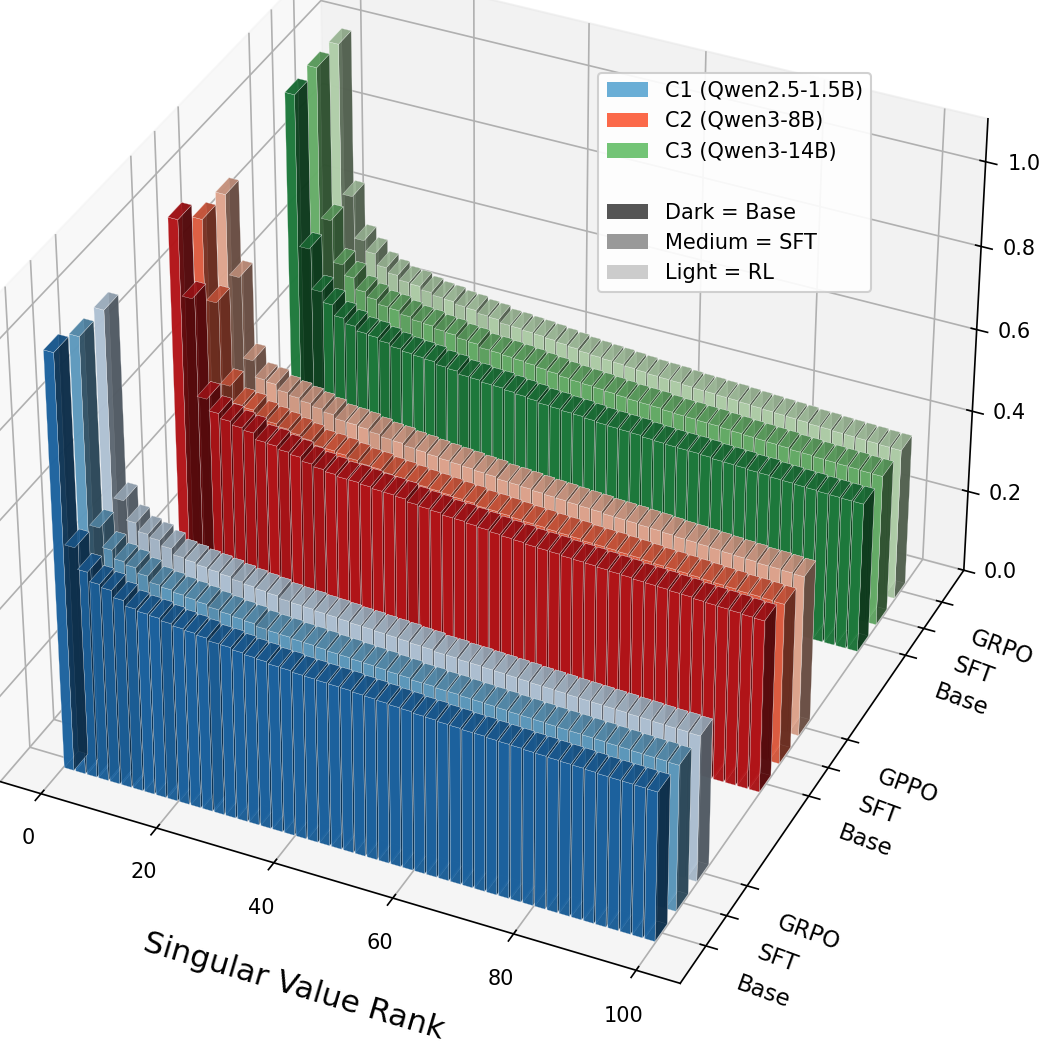}
    \caption{Top 100 singular values from base, SFT, RL.}
    \label{fig:spectrum_top100}
  \end{subfigure}
  \hfill
  \begin{subfigure}[b]{0.48\textwidth}
    \includegraphics[width=\textwidth]{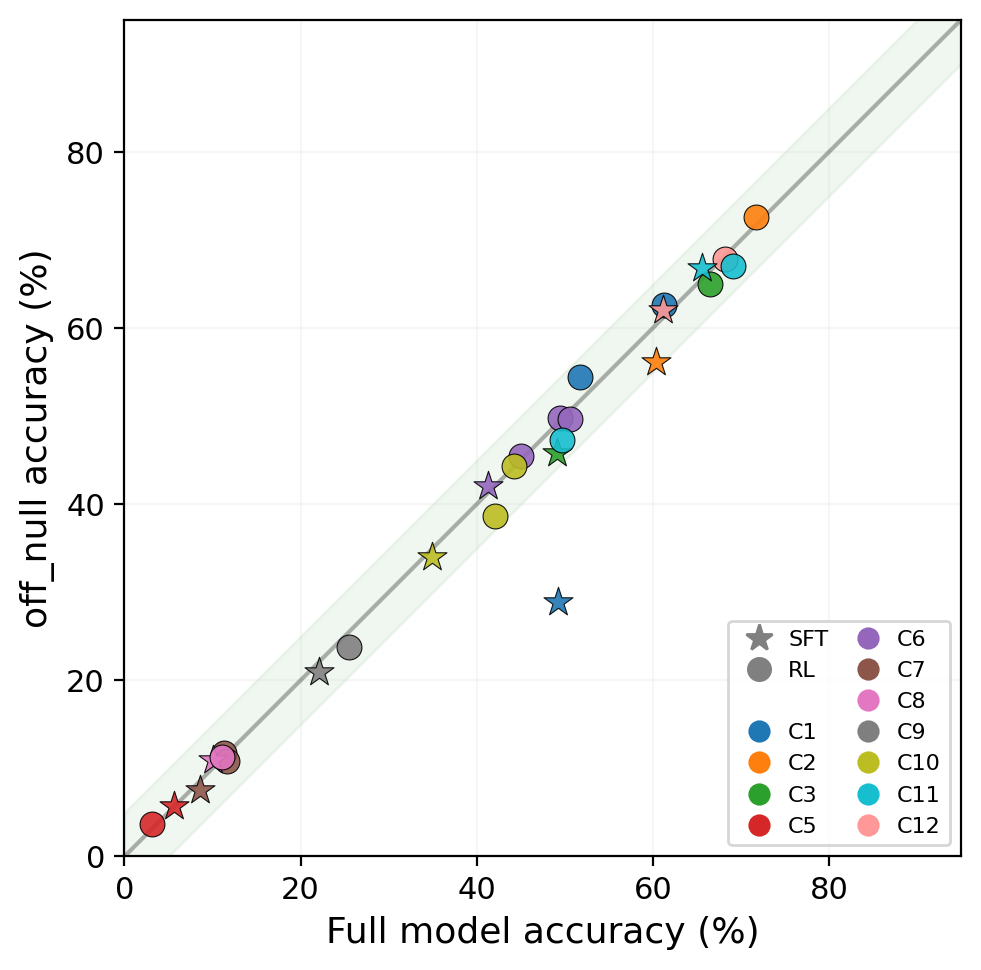}
    \caption{Removing the diagonal almost never hurts.}
  \label{fig:causal}
  \end{subfigure}
  \caption{
  Left: the leading singular values remain visually almost identical before and
  after post-training. We plot three chains with large spectral changes. Right: Across nearly all tested transitions, removing the diagonal reproduces the full post-trained model with trivial degradation with the exception of C1.}
  \label{fig:combined_singular_value}
\vspace{-0.2cm} 
\end{figure}

Prior work uses spectral structure to explain different aspects of post-training. \citet{zhu2025pathtakenrlvrprovably} finds that RL training is steered away from principal or diagonal directions, while \citet{he2026understandingposttrainingstructuralchanges} finds near uniform singular value scaling in post-training with coordinated rotations. Our post-hoc result is complementary to previous studies, and we find the singular value profile before and after post-training is extremely similar, and the leading singular values are almost unchanged (Figure~\ref{fig:combined_singular_value}a). Even when the update is relatively large (relative weights change values are in Table \ref{tab:spectral_invariance}), post-training rarely changes the pretrained spectrum.

This does not mean the diagonal update is always zero. Larger updates can place more energy in the diagonal channel as shown in Appendix \ref{app:energy-fractions}, and SFT often has larger relative updates than RL. We do not see $\dW_{\mathrm{diag}}$ in the model chains we study can cleanly separate SFT from RL, nor does it meaningfully reshape the pretrained spectrum. We find similar results as \citet{he2026understandingposttrainingstructuralchanges} that large $\dW_{\mathrm{diag}}$ shows a near uniform scale shift of $\Sig$. Figure~\ref{fig:spectrum_top100} shows the normalized spectra for the chains with the largest changes. We find the profiles are nearly identical. To measure quantitatively, we turn the $\Sig$ into vector and measure the cosine similarity between the pretrained and post-trained singular values. We find the spectral cosine exceeds 0.9999 for every transition. The spectrum profile is effectively unchanged. Detailed results are in Appendix~\ref{app:spectral_invariance}.
This suggests that the useful part of post-training may lie less in $\dW_{\mathrm{diag}}$ and more in how computation is rerouted within and beyond the pretrained subspace.


\subsection{Removing the Diagonal Usually Does Not Hurt}
\label{sec:causal_decomp}

The spectral analyses above are descriptive as they only show modest difference of singular-value profile, but they do not establish whether the diagonal component is functionally important.
We therefore test its causal contribution by removing $\Delta W_{\mathrm{diag}}$ from the post-trained model and reevaluating performance. 
Equivalently, it means keeping only $\dW_{\mathrm{off}} + \dW_{\mathrm{null}}$ on top of the base model. This intervention measures how much of the post-training improvement depends on the learned diagonal component. Figure~\ref{fig:causal} reports the average performance on our math-heavy evaluation suite, while Appendix~\ref{app:causal-eval} provides the complete benchmark-level results.
Across the 25 positive gain transitions defined in Section~\ref{sec:setup}, the model without $\dW_{\mathrm{diag}}$ stays within 4.3 points of the full post-trained model in 24 cases. The single clear exception is C1 distillation SFT, where removing the diagonal change causes a 20.3-point drop.

C1 is also an outlier in the geometry of the update. It distills the 671B DeepSeek-R1 into a 1.5B Qwen2.5-Math-1.5B base~\citep{Guo_2025}, making it the most aggressive distillation setting in our study. It has the highest diagonal energy among SFT transitions and one of the largest spectral perturbations (Appendix~\ref{app:spectral_invariance} and Appendix~\ref{app:energy-fractions}), and it is the only case where we see removing the $\dW_{\mathrm{diag}}$ causes a large behavioral failure. Inspection of the generated outputs suggests that the $\dW_{\mathrm{diag}}$  in C1 helps control the model's self-reflection behavior. After $\dW_{\mathrm{diag}}$ removal, the model frequently enters extended reflection loops, repeatedly changing direction without producing a final answer. The average number of ``Wait'' tokens rises from 46.4 to 171.7 per answer and model keeps changes direction without giving a final answer. This suggests that the diagonal change in C1 distillation SFT carries behaviorally important information. We treat it as a heavy distillation outlier rather than evidence that the diagonal is generally necessary, and other 24 cases do not exhibit a comparable dependence on $\Delta W_{\mathrm{diag}}$. Moreover, the subsequent DeepScaleR and Nemotron GRPO transitions initialized from the C1 distilled checkpoint do not show the same diagonal dependence.

\paragraph{Is it the removing size or the diagonal subspace?}
In most transitions, the $\dW_{\mathrm{diag}}$ contains only a small fraction of the total update energy. Consequently, good performance after $\dW_{\mathrm{diag}}$ removal could simply reflect that little update energy was removed, rather than that the diagonal direction is unusually unimportant. To distinguish component magnitude from component direction, we compare diagonal removal with a same-norm $\left|\Delta W_{\mathrm{diag}}\right|_F$ random-removal control. We test it on two chains, C1 and C2, since they carry substantial diagonal energy, whereas other transitions carry at most 0.011. The results are in Appendix \ref{app:rando} Table \ref{tab-remove-diag}.

The two chains are in opposite directions. In C2, removing the $\dW_{\mathrm{diag}}$ costs 4.3 points while removing the same energy in a random direction costs 15.2. It implies that removing the diagonal is cheaper than removing a random direction. C1 is the opposite, as expected, since we show in the paper that the diagonal is important for C1. In the remaining transitions, both diagonal removal and same-norm random removal have small effects within noise ($\pm 3\%$), so we cannot discriminate between them there.

These results distinguish diagonal \emph{change} from diagonal \emph{contribution}. A large diagonal component need not carry a large fraction of the performance gain, as shown by C2, while C1 demonstrates that the diagonal can matter under an exceptional heavy-distillation shift. Overall, however, diagonal reshaping is not the necessary carrier of most post-training gains in the transitions we study: removing it preserves nearly all of the full model's performance in 24 of 25 positive-gain cases. The remaining gains are retained primarily by off-diagonal mixing within the pretrained SVD core and null-space routing outside it, whose relative roles we examine next.



\section{Useful Changes Live in the Off-Diagonal and Null Space Terms}
\label{sec:interaction}

\begin{figure}[ht]
  \centering
  \includegraphics[width=\textwidth]{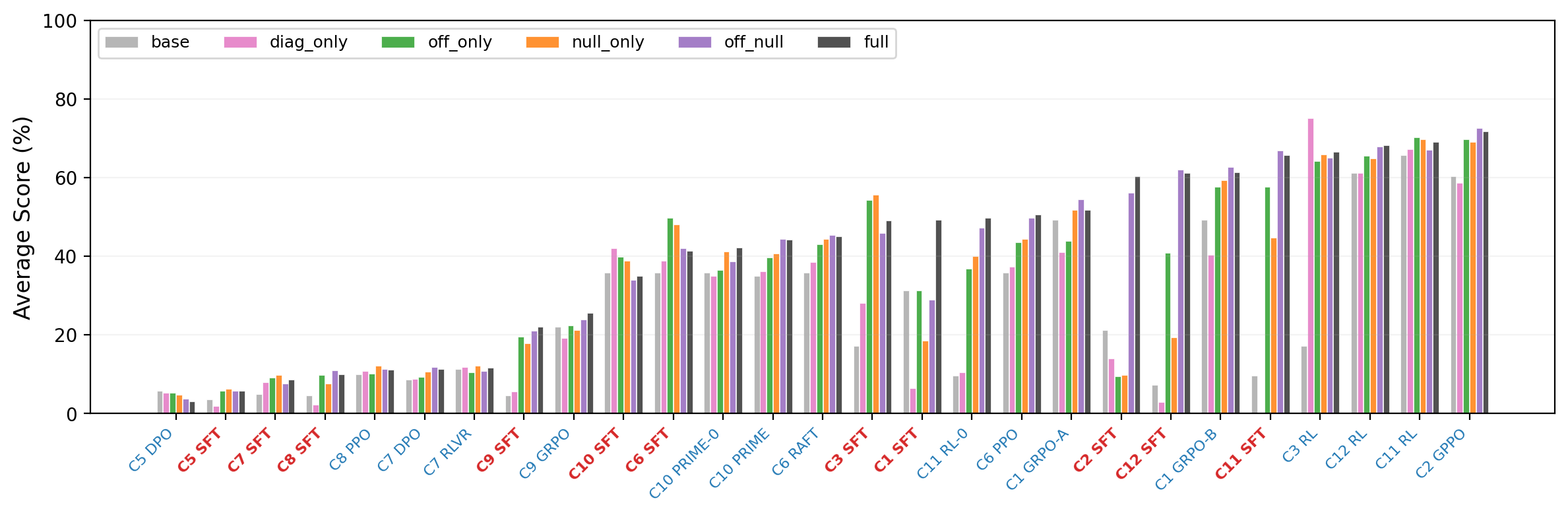}
  \caption{
    \textbf{Ablation scores relative to base and full model.}
    Each bar is one transition, sorted by full score.
    The gray bar marks the base model score; the black bar
    marks the full post-trained model score.
    Colored markers show the performance of each isolated component:
    diagonal only (pink),
    off-diagonal only (green),
    null space only (orange),
    and OffNull (purple). Red names correspond to SFT models, and blue names are RL models.
  }
  \label{fig:raw_scores}
\end{figure}
\vspace{-0.2cm}

GRRR leaves two candidates of performance gain after the diagonal is removed: off-diagonal updates ($\dW_{\mathrm{off}}$), which remix existing singular directions, and null space updates ($\dW_{\mathrm{null}}$), which route outside the pretrained frame. We ask two concrete questions: how much does each component recover on its own, and what changes when we combine them? We analyze all 28 causal decompositions. We follow Section~\ref{sec:setup}, and it leaves 25 informative transitions: 10 SFT and 15 RL. Figure~\ref{fig:raw_scores} plots the raw five benchmark scores for the full set.

\begin{figure}[ht]
  \vspace{-0.2cm}
    \centering
    \begin{subfigure}[b]{0.40\textwidth}
    \includegraphics[width=\textwidth]{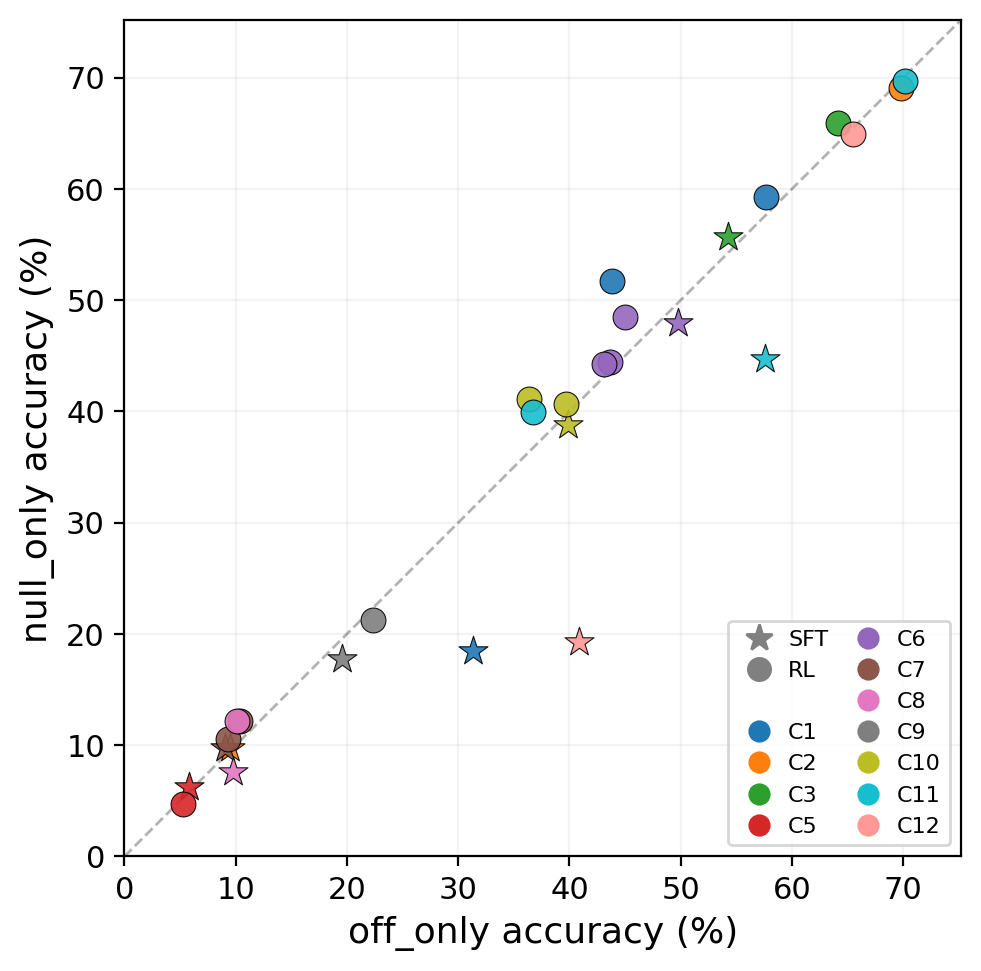}
      \caption{\textbf{Off vs null component performance.}
      Each point is one transition plotted by its off-diagonal performance against its null space performance in raw accuracy.
      }
      \label{fig:offnull_scatter}
    \end{subfigure}
    \hfill
    \begin{subfigure}[b]{0.41\textwidth}
      \includegraphics[width=\textwidth]{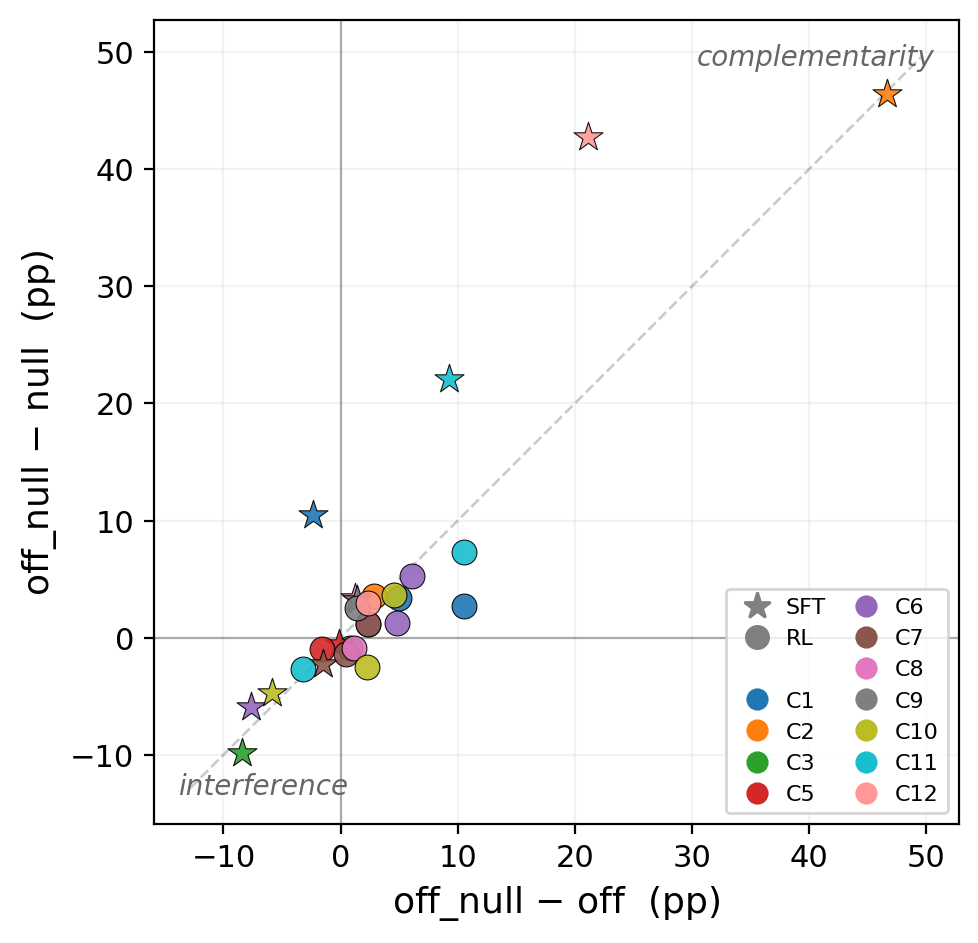}
      \caption{ \textbf{Interaction between off-diagonal and null space updates.} Positive values indicate complementarity. Negative values indicate interference.
      }
    \label{fig:interaction}
    \end{subfigure}
    \vspace{-0.2cm}
    \caption{Left: off vs null accuracy. Right: interaction of off and null}
    \label{fig:contribution}
  \vspace{-0.5cm}
  \end{figure}

\paragraph{Individual recovery.}
The OffNull model recovers nearly the full post-training gain in almost every informative case. We do not see a clean split by SFT/RL algorithm differences, and the more consistent pattern is redundancy versus asymmetry. RL transitions often recover well from either component alone, so off and null redundantly store updates. SFT transitions show wider variance and stronger asymmetry between the two components, with the dominant channel varying by chain. At the pooled level, SFT more often has one component clearly ahead of the other, while RL more often places the two components near parity. 
Figure~\ref{fig:offnull_scatter} shows this more directly: RL points cluster near the equal contribution diagonal, while SFT points are more spread out and often lean toward the off-diagonal component for performance gain.

\paragraph{Interaction.}
We next ask whether keeping both off and null terms adds anything beyond either component alone. Let ${\rm OffNull}$ denote the model that keeps the off and null terms together. We track the two signed quantities
\begin{equation}
  ({\rm OffNull} - {\rm off}) \quad \text{and} \quad ({\rm OffNull} - {\rm null}), 
\end{equation}
where positive values indicate complementarity and negative values indicate interference. Values near zero mean that the combination adds little over the stronger single component. Figure~\ref{fig:interaction} shows the result.
In RL, most values are small and slightly positive. Off and null carry largely redundant information, so combining them gives a modest but reliable gain. In SFT, the distribution is wider, more asymmetric, and sometimes negative. Some transitions benefit substantially from retaining both components, while in others an individual component can outperform their combination.
Once the diagonal is removed, the gain lives in rotation and routing. RL updates more often show redundancy between the two channels, whereas SFT updates exhibit greater asymmetry and stronger interaction. We next ask whether successive stages of post-training reuse the same update directions or move into different parts of the weight space.

\section{Sequential SFT and RL Have Nearly Orthogonal Rotations}
\label{sec:orthogonal}

We do not see the SFT and RL difference geometrically from the section above, but are there other differences? The GRRR framework allows us to compare the SFT and RL under the same base model coordinates. Modern pipelines usually follow a sequence with SFT first, then RL. This makes the geometry of the pipeline interesting to study. \citet{jin2025rlfinetuninghealsood} argue that RL softly realigns singular vectors that SFT had rotated. \citet{mukherjee2025reinforcement} find that RL touches only 5--30\% of parameters, which suggests limited overlap with SFT. GRRR allows us to ask directly whether the SFT and incremental RL weight updates are aligned, opposed, or nearly orthogonal.

Because both SFT and RL updates are projected into the same pretrained SVD basis, we can compare them directly. We study the standard sequential setting (base $\to$ SFT $\to$ RL) and decompose the cumulative update
as
$\Delta W_{\text{cum}} = \Delta W_{\text{SFT}} + \Delta W_{\text{incr}}$
where $\Delta W_{\text{SFT}} = W_{\text{SFT}} - W_{\text{base}}$ and
$\Delta W_{\text{incr}} = W_{\text{RL}} - W_{\text{SFT}}$.

\begin{figure}
  \centering
  \includegraphics[width=\textwidth]{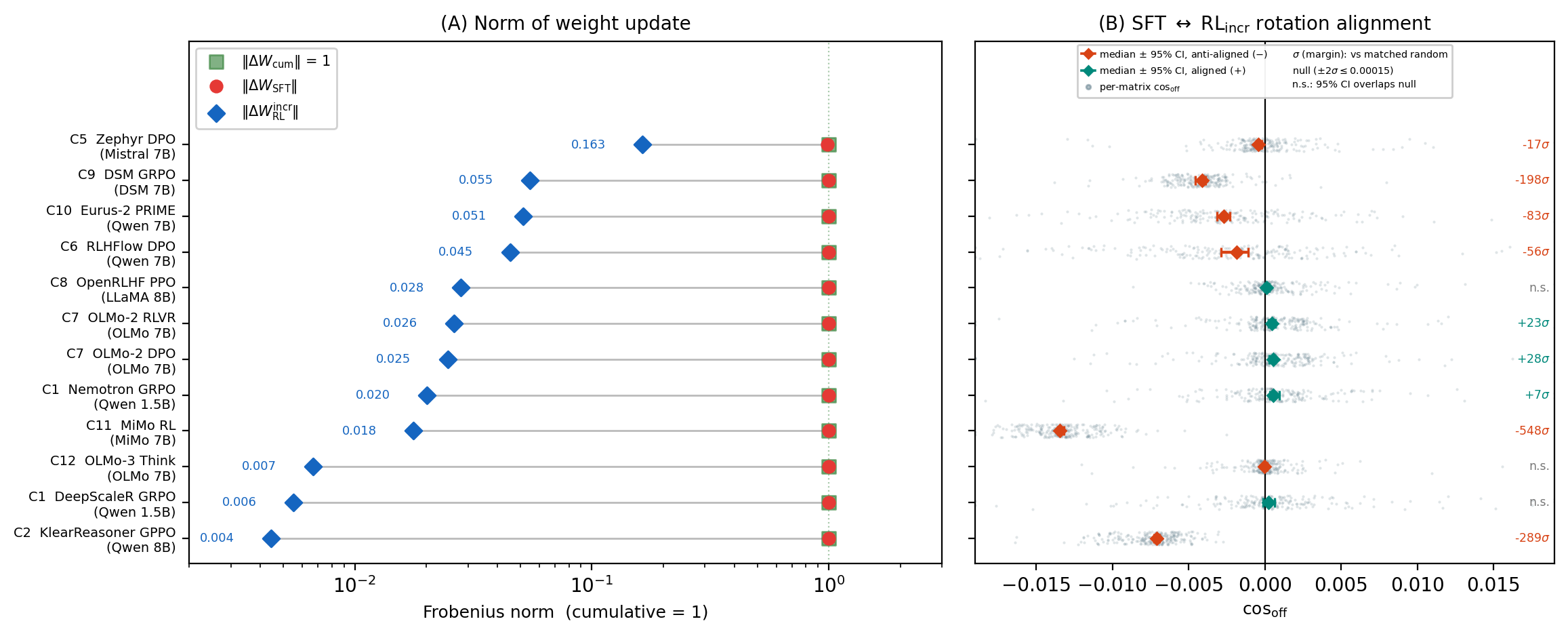}
  \caption{
    \textbf{(A)}~Normalized Frobenius norms (log scale) of the SFT,
    incremental RL, and cumulative updates. SFT dominates in all chains.
    \textbf{(B)}~ Off-diagonal cosine similarity between SFT and
    incremental RL rotations. Diamonds show the median over matrices and bars show bootstrap 95\% confidence intervals. Significance is measured against a matrix- and chain-matched isotropic null for the same median statistic.}
  \label{fig:cosine}
\end{figure}

\paragraph{SFT dominates update magnitude.}
Figure~\ref{fig:cosine}A shows the Frobenius norm of each component, normalized so that $\|\Delta W_{\text{cum}}\| = 1$. Across all chains, SFT accounts for nearly the entire update magnitude. The incremental RL step is one to two orders of magnitude smaller ($\approx 0.004$--$0.17\times$). In raw size, the pipeline is dominated by SFT.

\vspace{-0.2cm}

\paragraph{RL Rotations Are Almost Orthogonal to SFT Rotations}
Magnitude does not determine direction, so we compare the off-diagonal components of the two updates. The rotation are the off-diagonal P-matrix in the pretrained SVD frame. We compute:
$
  \cos_{\text{off}}
    = \frac{
        \langle P_{\text{SFT}}^{\text{off}},\;
                P_{\text{incr}}^{\text{off}} \rangle_F
      }{
        \lVert P_{\text{SFT}}^{\text{off}} \rVert_F \;\cdot\;
        \lVert P_{\text{incr}}^{\text{off}} \rVert_F
      }$ as the similarity measure of the post-trained off-diagonal changes.
Figure~\ref{fig:cosine}B shows that the median cosine remains small in absolute value. However, near-zero cosine alone is not sufficient evidence because random vectors are already nearly orthogonal in high-dimensional spaces. The key is whether the cosine is smaller than chance. We therefore compare each chain against an isotropic null matched to its own matrix shapes, and to the matrices statistic. We plot the result in the Figure~\ref{fig:cosine}B. Because cosine is scale invariant, matching the matrix dimensions and aggregation statistic determines the appropriate null.

Under this baseline, 9 of the 12 sequential transitions differ significantly from chance. 6 are significantly negative, 3 are weakly positive, and 3 are statistically indistinguishable from the null. The negative pattern is also consistent across matrices. The negative values are not explained by high-dimensional noise. Their small absolute magnitude indicates partial reversal rather than strong opposition. This observation is consistent with \citeauthor{jin2025rlfinetuninghealsood}'s observation that RL can partially reverse rotations induced by SFT. The more informative comparison is with SFT and RL updates trained independently from the same base model. As shown in Appendix \ref{sec:cross_model} Figure ~\ref{fig:cross_model} B, independent SFT and RL updates are substantially more aligned when we compare their cosine values, whereas incremental RL following SFT is nearly orthogonal or weakly anti-aligned. More discussions are in Appendix \ref{sec:cross_model} and comparing sequential and independent trained SFT/RL is more informative.

\section{Discussion and Limitations}
\label{sec:discussion}

Our study is limited to decoder LLMs, and our strongest performance claims come from a five-benchmark average dominated by math reasoning. We study fully trained checkpoints and the final weight difference $\Delta W$, not training-time interventions or optimization trajectories. Our causal claims are about post-hoc ablations of the final $\Delta W$. We also study only the linear weight matrices; activations, normalization layers, optimizers, and training dynamics are outside the scope of this paper.

The following are three observations from the weight space. (1) Independent SVDs remain almost unchanged after post-training (Appendix~\ref{app:spectral_invariance}), effective rank of matrices are stable (Appendix~\ref{app:rank-invariance}), and removing the diagonal usually does not hurt performance. Together, these results suggest that post-training usually remains near an approximately isospectral neighborhood of the pretrained model. The main functional change is not a reconstruction of the singular spectrum, but a reorganization of how the pretrained weight space is used, specifically in the off-diagonal and null terms.
(2) At the matrix level, the diagonal changes the strength of matched singular vector channels. Off-diagonal terms remix those singular vector channels. Null terms change what the matrix can see and write, and they let a layer to read/write hidden state directions outside its pretrained input subspace. It allows the layer to read/write features previously ignored. 
(3) The sequential SFT then RL pipeline adds another constraint. We see a contrast between the sequential and the independent SFT/RL. We believe the algorithm label alone is insufficient, and controlled data and compute are more important as the controlled C3 SFT and RL runs are nearly identical in both direction and magnitude in Figure \ref{fig:cross_model}. 
Sequential SFT/RL show a more interesting relationship in weight space. When RL starts after SFT, it tends to move into directions that the preceding update did not use and sometimes partially reverses it, rather than simply amplifying the same rotation. 


\section{Conclusion}
\label{sec:conclusion}

We introduce GRRR, a simple decomposition of post-training weight changes in the pretrained SVD frame into diagonal reshaping, off-diagonal rotation, and null space routing. This decomposition extends SVD-based weight analysis from description to function. GRRR reconstructs models from each component and directly tests which parts of the final weight difference carry the resulting behavior.
Across the decoder LLM settings we study, the main result is that the diagonal spectral reshaping is usually not the necessary carrier of post-training gains. Most ordinary post-training gains do not require rebuilding spectral values except heavy distillation.
The useful change is retained mainly through two forms of reorganization -- singular vectors and null space. 
Together, these results suggest a distinction between the roles of pretraining and post-training: pretraining constructs the model's weight-space structure, while post-training primarily reorganizes how that structure is used. This view opens several directions. Orthogonal or weakly interfering update subspaces may enable more reliable model merging and weight arithmetic; separating changes in the left and right singular directions may clarify how post-training modifies what a layer reads and writes; and finer analysis across attention, MLP, and other matrix types may reveal where different forms of routing occur. GRRR provides a framework for studying post-training as the organization of pretrained capacity rather than its reconstruction.



\section*{Acknowledgments}
\label{sec:acknowledgements}
The work is supported by the National Science Foundation (NSF) through Awards \#DMR-2433348 (AI Research Institutes),  \#2131186 (CISE-MSI),  \#1827505 (PFI), and the US Air Force Office of Scientific Research (AFOSR) via Award \#FA9550-21-1-0082. The work is also supported by a CCNY College-wide Research Vision (CRV) Fund (2022-2025) and the Google CyberNYC Initiative (2025-2030). This work used Google Cloud through the CloudBank project, which is supported by NSF Award \#1925001.
We specifically thank Dr. Ping Ji's generous support for computing resources. 

\clearpage

\bibliography{colm2026_conference}
\bibliographystyle{plainnat}

\appendix


\clearpage

\section{Disclosure}
\subsection{LLM Usage Disclosure}
We used large language models as research assistants throughout the project. Specifically, LLMs were used to support code implementation, assist in refining hypotheses and experimental design, aid in literature discovery, and help improve writing clarity and organization.
We used OpenAI Codex and Anthropic Claude Code to assist with research engineering tasks, including implementing and debugging experimental code, launching and monitoring experiments, analyzing intermediate outputs, and maintaining experiment records. We additionally used the \texttt{delta-research} framework\footnote{\url{https://github.com/user074/delta-research}} to organize iterative hypothesis testing and experiment execution. 

These tools substantially accelerated the research process. The core research ideas, problem formulation, and overall methodology were initiated and guided by the authors. All generated code, experimental results, and analyses were carefully reviewed, validated, and iterated upon by the authors. The authors take full responsibility for the correctness, originality, and integrity of all content in this paper.

\subsection{Reproducibility Disclosure}
We include code, scripts, and instructions necessary to reproduce our results.

The codebase includes data processing pipelines, model loading, training and evaluation scripts, and analysis code for reported metrics and figures. In particular, we provide scripts to recompute all weight-space analyses, including SVD-based decompositions and alignment measures, directly from model checkpoints.

\section{Complete Transitions Table}
\label{app:transitions}

Table~\ref{tab:chains} lists the model chains analyzed in this study. Each
chain starts from a pretrained base model and applies one or more post-training
stages. The current inventory contains 12 chains labeled C1 through C12.
Table~\ref{tab:transitions} gives the complete inventory of all 29
incremental transitions. Later appendices additionally analyze cumulative
base to post transitions derived from the same chains, which is why those
sections report larger transition counts.

\begin{table}[ht]
\centering
\small
\caption{Chain inventory. C6 and C10 share the same pretrained base
(Qwen2.5-Math-7B) but use different post-training pipelines.
$\dagger$\,indicates training directly from the pretrained base rather than
from the preceding SFT checkpoint.}
\label{tab:chains}
\resizebox{\textwidth}{!}{%
\begin{tabular}{clrl}
\toprule
Chain & Base Model (HF identifier) & Size & Pipeline \\
\midrule
C1  & Qwen/Qwen2.5-Math-1.5B \citep{yang2024qwen25mathtechnicalreportmathematical}         & 1.5B & SFT $\to$ GRPO ($\times$2) \\
C2  & Qwen/Qwen3-8B-Base \citep{yang2025qwen3technicalreport}             & 8B   & SFT $\to$ GPPO \\
C3  & Qwen/Qwen3-14B    \citep{yang2025qwen3technicalreport}              & 14B  & SFT $\to$ GRPO \\
C4  & Qwen/Qwen3-8B-Base  \citep{yang2025qwen3technicalreport}             & 8B   & GRPO (direct) \\
C5  & mistralai/Mistral-7B-v0.1 \citep{jiang2023mistral7b}      & 7B   & SFT $\to$ DPO \\
C6  & Qwen/Qwen2.5-Math-7B   \citep{yang2024qwen25mathtechnicalreportmathematical}         & 7B   & SFT $\to$ DPO / PPO$^\dagger$ / RAFT$^\dagger$ \\
C7  & allenai/OLMo-2-1124-7B  \citep{olmo20252olmo2furious}        & 7B   & SFT $\to$ DPO $\to$ RLVR \\
C8 & meta-llama/Meta-Llama-3-8B  \citep{grattafiori2024llama3herdmodels}    & 8B   & SFT $\to$ PPO \\
C9 & deepseek-ai/deepseek-math-7b-base \citep{shao2024deepseekmathpushinglimitsmathematical}& 7B & SFT $\to$ GRPO \\
C10 & Qwen/Qwen2.5-Math-7B     \citep{yang2024qwen25mathtechnicalreportmathematical}       & 7B   & SFT $\to$ PRIME / PRIME$^\dagger$ \\
C11 & XiaomiMiMo/MiMo-7B-Base  \citep{xiaomi2025mimounlockingreasoningpotential}       & 7B   & SFT $\to$ RL / RL$^\dagger$ \\
C12 & allenai/OLMo-3-1025-7B   \citep{olmo2025olmo3}       & 7B   & SFT $\to$ DPO{+}OlmoRL \\
\bottomrule
\end{tabular}
}
\end{table}

\begin{table*}[ht]
\centering
\scriptsize
\caption{Complete inventory of incremental transitions. Each row is one
training step producing the listed model. ``From'' gives the full name of the
immediate predecessor model. $\dagger$\,marks custom checkpoints not publicly
released on HuggingFace. Domain indicates the training objective:
\textbf{Math} = mathematical reasoning, \textbf{General} = general
instruction following, \textbf{Reasoning} = broad reasoning / thinking,
\textbf{Agent} = agentic tool use, \textbf{Metacog.} = metacognitive
self-reflection.}
\label{tab:transitions}
\setlength{\tabcolsep}{3.5pt}
\resizebox{\textwidth}{!}{%
\begin{tabular}{cllrlll}
\toprule
Chain & Family & Trained Model & Size & From & Method & Domain \\
\midrule
C1  & Qwen 2.5 & DeepSeek-R1-Distill-Qwen-1.5B \citep{Guo_2025}  & 1.5B & Qwen2.5-Math-1.5B                & SFT  & Reasoning \\
C1  & Qwen 2.5 & DeepScaleR-1.5B-Preview \citep{deepscaler2025}         & 1.5B & DeepSeek-R1-Distill-Qwen-1.5B    & GRPO & Math \\
C1  & Qwen 2.5 & Nemotron-Reasoning-Qwen-1.5B \citep{liu2025prorlprolongedreinforcementlearning}     & 1.5B & DeepSeek-R1-Distill-Qwen-1.5B    & GRPO & Math \\
\addlinespace
C2  & Qwen 3   & Klear-Reasoner-8B-SFT \citep{su2026klearreasoneradvancingreasoningcapability}           & 8B   & Qwen3-8B-Base                    & SFT  & Reasoning \\
C2  & Qwen 3   & Klear-Reasoner-8B  \citep{su2026klearreasoneradvancingreasoningcapability}               & 8B   & Klear-Reasoner-8B-SFT            & GPPO & Reasoning \\
\addlinespace
C3  & Qwen 3   & UniReason-Qwen3-14B-SFT  \citep{huan2025doesmathreasoningimprove}        & 14B  & Qwen3-14B                        & SFT  & Reasoning \\
C3  & Qwen 3   & UniReason-Qwen3-14B-RL \citep{huan2025doesmathreasoningimprove}          & 14B  & Qwen3-14B          & GRPO & Reasoning \\
\addlinespace
C4  & Qwen 3   & SkyRL-Agent-WebResearch-8B \citep{cao2025skyrlagentefficientrltraining}       & 8B   & Qwen3-8B-Base                    & GRPO & Agent \\
\addlinespace
C5  & Mistral  & mistral-7b-sft-beta   \citep{tunstall2023zephyrdirectdistillationlm}            & 7B   & Mistral-7B-v0.1                  & SFT  & General \\
C5  & Mistral  & zephyr-7b-beta \citep{tunstall2023zephyrdirectdistillationlm}                   & 7B   & mistral-7b-sft-beta              & DPO  & General \\

\addlinespace
C6  & Qwen 2.5 & Qwen2.5-7B-SFT \citep{zhangonline}                  & 7B   & Qwen2.5-Math-7B                  & SFT  & Math \\
C6  & Qwen 2.5 & Qwen2.5-7B-DPO   \citep{zhangonline}                & 7B   & Qwen2.5-7B-SFT                   & DPO  & Math \\
C6  & Qwen 2.5 & Qwen2.5-7B-PPO-Zero   \citep{zhangonline}           & 7B   & Qwen2.5-Math-7B                  & PPO  & Math \\
C6  & Qwen 2.5 & Qwen2.5-7B-RAFT-Zero    \citep{zhangonline}         & 7B   & Qwen2.5-Math-7B                  & RAFT & Math \\
\addlinespace
C7  & OLMo 2   & OLMo-2-1124-7B-SFT   \citep{olmo20252olmo2furious}            & 7B   & OLMo-2-1124-7B                   & SFT  & General \\
C7  & OLMo 2   & OLMo-2-1124-7B-DPO      \citep{olmo20252olmo2furious}          & 7B   & OLMo-2-1124-7B-SFT               & DPO  & General \\
C7  & OLMo 2   & OLMo-2-1124-7B-Instruct    \citep{olmo20252olmo2furious}        & 7B   & OLMo-2-1124-7B-DPO               & RLVR & General \\
\addlinespace
C8 & LLaMA 3  & Llama-3-8b-sft-mixture  \citep{dong2024rlhfworkflowrewardmodeling}          & 8B   & Meta-Llama-3-8B                  & SFT  & General \\
C8 & LLaMA 3  & Llama-3-8b-rlhf-100k   \citep{dong2024rlhfworkflowrewardmodeling}           & 8B   & Llama-3-8b-sft-mixture           & PPO  & General \\
\addlinespace
C9 & DeepSeek & deepseek-math-7b-instruct  \citep{shao2024deepseekmathpushinglimitsmathematical}        & 7B   & deepseek-math-7b-base            & SFT  & Math \\
C9 & DeepSeek & deepseek-math-7b-rl    \citep{shao2024deepseekmathpushinglimitsmathematical}             & 7B   & deepseek-math-7b-instruct        & GRPO & Math \\
\addlinespace
C10 & Qwen 2.5 & Eurus-2-7B-SFT \citep{cui2025processreinforcementimplicitrewards}                   & 7B   & Qwen2.5-Math-7B                  & SFT  & Math \\
C10 & Qwen 2.5 & Eurus-2-7B-PRIME    \citep{cui2025processreinforcementimplicitrewards}              & 7B   & Eurus-2-7B-SFT                   & PRIME & Math \\
C10 & Qwen 2.5 & Eurus-2-7B-PRIME-Zero    \citep{cui2025processreinforcementimplicitrewards}         & 7B   & Qwen2.5-Math-7B                  & PRIME & Math \\
\addlinespace
C11 & MiMo     & MiMo-7B-SFT     \citep{xiaomi2025mimounlockingreasoningpotential}                  & 7B   & MiMo-7B-Base                     & SFT  & Reasoning \\
C11 & MiMo     & MiMo-7B-RL       \citep{xiaomi2025mimounlockingreasoningpotential}                 & 7B   & MiMo-7B-SFT                      & RL   & Reasoning \\
C11 & MiMo     & MiMo-7B-RL-Zero     \citep{xiaomi2025mimounlockingreasoningpotential}              & 7B   & MiMo-7B-Base                     & RL   & Reasoning \\
\addlinespace
C12 & OLMo 3   & OLMo-3-7B-Think-SFT    \citep{olmo2025olmo3}          & 7B   & OLMo-3-1025-7B                   & SFT  & Reasoning \\
C12 & OLMo 3   & OLMo-3-7B-Think    \citep{olmo2025olmo3}               & 7B   & OLMo-3-7B-Think-SFT              & DPO{+}OlmoRL & Reasoning \\
\bottomrule
\end{tabular}
}
\end{table*}

\section{Spectral Invariance of post-training}
\label{app:spectral_invariance}

post-training leaves the singular value spectrum of weight matrices almost unchanged, even when the update is large. This appendix reports results for all 29 transitions in our inventory: 13 post-SFT follow-up transitions, 6 from-base RL transitions, and 11 base$\to$SFT transitions.

\subsection{Metrics}

For each checkpoint pair $(W_\text{pre}, W_\text{post})$, we summarize the update size by
\begin{equation}
    \rupd = \frac{\|W_\text{post} - W_\text{pre}\|_F}{\|W_\text{pre}\|_F},
\end{equation}
where the Frobenius norms aggregate all seven decomposed matrices across all layers in the transition. Table~\ref{tab:spectral_invariance} is sorted by $\rupd$ within each block.

At the matrix level, we then compute singular values with \texttt{svdvals} and report two complementary measures of spectral change.

\paragraph{Singular Value (SV) Cosine.} 
For a matrix $W\in\mathbb{R}^{m\times n}$, let $k=\min(m,n)$ and define its ordered singular-value vectors as
\begin{equation}
\boldsymbol{\sigma}(W)
=
\bigl(\sigma_1(W),\sigma_2(W),\ldots,\sigma_k(W)\bigr),
\qquad
\sigma_1(W)\geq \sigma_2(W)\geq\cdots\geq\sigma_k(W)\geq 0.
\end{equation}
The entries of $\boldsymbol{\sigma}(W)$ are scalar singular values. We define it as singular value vectors so we can compare across the matrices of their singular value spectrum similarity. They should not be confused with the left and right singular vectors contained in $U$ and $V$.

The cosine similarity between the pre and post-training singular value vectors:
\begin{equation}
    \text{SV cosine} = \cos(\boldsymbol{\sigma}_\text{pre},\, \boldsymbol{\sigma}_\text{post})
    = \frac{\boldsymbol{\sigma}_\text{pre} \cdot \boldsymbol{\sigma}_\text{post}}
           {\|\boldsymbol{\sigma}_\text{pre}\|\,\|\boldsymbol{\sigma}_\text{post}\|}.
\end{equation}

SV cosine $= 1.0$ means the spectral shape is perfectly preserved. Because singular values follow a steep power law, $\sigma_1$ dominates the dot product, so this metric is most sensitive to changes in the largest singular values.

\paragraph{Maximum Relative Perturbation.} For one matrix, the maximum relative perturbation is
\begin{equation}
    \delta_{\max}
    = \max_i \frac{|\sigma_i^\text{post} - \sigma_i^\text{pre}|}{\sigma_1^\text{pre}}.
\end{equation}
This normalizes each singular value change by the spectral scale $\sigma_1^\text{pre}$ and gives a scale invariant worst case measure. We use SV cosine as a global summary and $\delta_{\max}$ as the stricter componentwise check.

\paragraph{Aggregation.} Each transition contains hundreds of weight matrices (7 module types across all layers). For each transition we report the mean and standard deviation of both SV cosine and $\delta_{\max}$ across matrices.

\subsection{Results}

Table~\ref{tab:spectral_invariance} reports results for all 29 transitions, with each block sorted by $\rupd$.

\begin{table}[t]
  \centering
  \caption{%
    Spectral invariance of weight matrices across 29 transitions: incremental post-SFT updates,
    from-base RL, and cumulative SFT, sorted by $\rupd$ within each group.
    SV cosine and $\delta_{\max}$ report mean $\pm$ std across all weight
    matrices in the transition.
  }
  \label{tab:spectral_invariance}
  \small
  \begin{tabular}{llccc}
  \toprule
  Chain & Transition & $\rupd$ & SV cos & $\delta_{\max}$ \\
  \midrule
  \multicolumn{5}{l}{\textit{Post-SFT follow-up transitions (12 transitions)}} \\[2pt]
  C6 (Qwen2.5-7B) & SFT$\to$DPO & 0.0003 & 1.000000{\tiny$\pm$0} & 0.01{\tiny$\pm$0.01}\% \\
  C8 (LLaMA-3-8B) & SFT$\to$PPO & 0.0010 & 1.000000{\tiny$\pm$0} & 0.01{\tiny$\pm$0.01}\% \\
  C1 (Qwen2.5-1.5B) & SFT$\to$GRPO-A & 0.0010 & 1.000000{\tiny$\pm$0} & 0.01{\tiny$\pm$0.01}\% \\
  C10 (Eurus-2-7B) & SFT$\to$PRIME & 0.0010 & 1.000000{\tiny$\pm$0} & 0.02{\tiny$\pm$0.01}\% \\
  C2 (Qwen3-8B) & SFT$\to$GPPO & 0.0020 & 1.000000{\tiny$\pm$0} & 0.02{\tiny$\pm$0.01}\% \\
  C9 (DSMath-7B) & SFT$\to$RL & 0.0020 & 1.000000{\tiny$\pm$0} & 0.02{\tiny$\pm$0.01}\% \\
  C12 (OLMo-3-7B) & SFT$\to$RL & 0.0020 & 1.000000{\tiny$\pm$0} & 0.02{\tiny$\pm$0.01}\% \\
  C7 (OLMo-2-7B) & SFT$\to$DPO & 0.0020 & 1.000000{\tiny$\pm$0} & 0.02{\tiny$\pm$0.04}\% \\
  C7 (OLMo-2-7B) & SFT$\to$Instruct & 0.0020 & 1.000000{\tiny$\pm$0} & 0.03{\tiny$\pm$0.04}\% \\
  C11 (MiMo-7B) & SFT$\to$RL & 0.0030 & 1.000000{\tiny$\pm$0} & 0.03{\tiny$\pm$0.01}\% \\
  C1 (Qwen2.5-1.5B) & SFT$\to$GRPO-B & 0.0050 & 1.000000{\tiny$\pm$0} & 0.02{\tiny$\pm$0.02}\% \\
  C5 (Mistral-7B) & SFT$\to$DPO & 0.0070 & 1.000000{\tiny$\pm$0} & 0.03{\tiny$\pm$0.02}\% \\
  \midrule
  \multicolumn{5}{l}{\textit{From-base RL: base $\to$ RL (5 transitions)}} \\[2pt]
  C6 (Qwen2.5-7B) & base$\to$PPO & 0.0010 & 1.000000{\tiny$\pm$0} & 0.01{\tiny$\pm$0.01}\% \\
  C10 (Qwen2.5-7B) & base$\to$PRIME-Zero & 0.0010 & 1.000000{\tiny$\pm$0} & 0.01{\tiny$\pm$0.01}\% \\
  C11 (MiMo-7B) & base$\to$RL-Zero & 0.0020 & 1.000000{\tiny$\pm$0} & 0.02{\tiny$\pm$0.01}\% \\
  C6 (Qwen2.5-7B) & base$\to$RAFT & 0.0050 & 1.000000{\tiny$\pm$0} & 0.02{\tiny$\pm$0.01}\% \\
  C4 (Qwen3-8B) & base$\to$GRPO & 0.0810 & 0.999999{\tiny$\pm$0.000002} & 1.29{\tiny$\pm$0.29}\% \\
  C3 (Qwen3-14B) & base$\to$GRPO & 0.0825 & 0.999999{\tiny$\pm$0.000002} & 1.29{\tiny$\pm$0.32}\% \\  \midrule
  \multicolumn{5}{l}{\textit{Cumulative SFT: base $\to$ SFT (11 transitions)}} \\[2pt]
  C6 (Qwen2.5-7B) & base$\to$SFT & 0.006 & 1.000000{\tiny$\pm$0} & 0.04{\tiny$\pm$0.04}\% \\
  C8 (LLaMA-3-8B) & base$\to$SFT & 0.019 & 1.000000{\tiny$\pm$0} & 0.08{\tiny$\pm$0.07}\% \\
  C10 (Eurus-2-7B) & base$\to$SFT & 0.024 & 1.000000{\tiny$\pm$0} & 0.11{\tiny$\pm$0.12}\% \\
  C9 (DSMath-7B) & base$\to$SFT & 0.030 & 1.000000{\tiny$\pm$0} & 0.24{\tiny$\pm$0.08}\% \\
  C5 (Mistral-7B) & base$\to$SFT & 0.038 & 1.000000{\tiny$\pm$0.000001} & 0.25{\tiny$\pm$0.23}\% \\
  C7 (OLMo-2-7B) & base$\to$SFT & 0.073 & 0.999998{\tiny$\pm$0.000004} & 0.62{\tiny$\pm$0.60}\% \\
  C3 (Qwen3-14B) & base$\to$SFT & 0.084 & 0.999999{\tiny$\pm$0.000002} & 1.29{\tiny$\pm$0.31}\% \\
  C11 (MiMo-7B) & base$\to$SFT & 0.189 & 0.999953{\tiny$\pm$0.000075} & 1.60{\tiny$\pm$0.75}\% \\
  C1 (Qwen2.5-1.5B) & base$\to$SFT & 0.192 & 0.999979{\tiny$\pm$0.000037} & 7.92{\tiny$\pm$11.45}\% \\
  C12 (OLMo-3-7B) & base$\to$SFT & 0.248 & 0.999976{\tiny$\pm$0.000033} & 5.05{\tiny$\pm$2.99}\% \\
  C2 (Qwen3-8B) & base$\to$SFT & 0.290 & 0.999930{\tiny$\pm$0.000114} & 8.63{\tiny$\pm$2.52}\% \\
  \bottomrule
  \end{tabular}
  \end{table}

\paragraph{Post-SFT follow-up transitions.}
All 12 post-SFT follow-up transitions achieve SV cosine $= 1.000000$ to 6 decimal places, with $\delta_{\max} < 0.04\%$. These updates are effectively isospectral. They redistribute weight energy across directions without meaningfully rescaling singular values.

\paragraph{Cumulative SFT.}
SV cosine is at least 0.9999 across all 11 transitions, including the most aggressive one, C2, where $\rupd = 0.29$ and SV cosine is still 0.99993. The $\delta_{\max}$ metric shows more variation with training intensity. Lighter SFT runs with $\rupd < 0.04$ perturb singular values by less than $0.3\%$ of $\sigma_1$, while the most aggressive SFT reaches $8.6\%$ in C2. Even there, no individual singular value moves by more than $8.6\%$ of the matrix's spectral scale.

\paragraph{From-base RL.}
The four direct math-RL branches, PPO-Zero, RAFT-Zero, PRIME-Zero, and RL-Zero, also show SV cosine $= 1.000000$ and $\delta_{\max} < 0.04\%$. RL alone is therefore isospectral in these runs as well. The only larger from-base RL case is C4 direct GRPO: SV cosine is still 0.999999, but $\delta_{\max}$ rises to 1.29\%.

\subsection{Discussion}

\paragraph{post-training Is Almost Isospectral}
Both SFT and RL preserve the singular value spectrum with high precision, even when they produce functionally different models. The clearest case is aggressive SFT: C2 base$\to$SFT has $\rupd = 0.29$ yet SV cosine $= 0.99993$. The weight matrix moves a lot, but the global spectral shape hardly changes. post-training mainly redistributes energy across directions. It changes which features are amplified and adds out-of-frame components through the null term. It does not substantially rescale existing singular values. This matches the main text, where the diagonal term is the least important component in the causal tests.



\section{Effective Rank Is Invariant Under post-training}
\label{app:rank-invariance}

The null component $\dW_{\mathrm{null}}$ lies outside the pretrained singular vector frame
$\mathcal{S} = \{\U_0 A \V_0^\top : A \in \R^{r \times r}\}$ and can therefore let a layer read from directions outside $\col(\V_0)$ or write into directions outside $\col(\U_0)$. This raises a natural question: does post-training increase the effective dimensionality of the weight matrix?

Algebraic rank is not informative here. At floating point precision, these matrices are already numerically full rank, and tiny perturbations in the spectral tail can move many near-zero singular values across a machine-epsilon threshold without reflecting a meaningful structural change. We therefore focus on \emph{effective rank}, namely how much of the spectral mass is concentrated in the dominant modes. In our data the stable rank of pretrained weights is much smaller than algebraic rank, typically between 109 and 184, so it can in principle change if post-training creates genuinely new dominant directions.

\paragraph{Metrics.}
We compute six rank metrics for $\W_0$, $\W_1 = \W_0 + \dW$, and $\dW$ across all matrices from 42 transition instances over 12 chains and 8 architecture families (1.5B--14B parameters). These are stable rank, $\varepsilon$-rank at $\varepsilon \in \{0.01, 0.001\}$, entropy erank, and the energy capture ranks $k_{50}$ and $k_{90}$. We also define the \textbf{rank additivity ratio}
\[
\alpha = (\mathrm{SR}(\W_1) - \mathrm{SR}(\W_0))\,/\,\mathrm{SR}(\dW),
\]
which equals 1 if the update's rank adds to the weight's rank and 0 if it is absorbed.

\paragraph{Per-transition detail.}
Table~\ref{tab:all-transitions} reports stable rank and additivity for all 42 transition instances. Two points stand out. First, rank is not additive. $\mathrm{SR}(\dW)$ ranges from 6.6 to 382, yet the median additivity ratio is $\alpha = -0.00002$. Even C11 MiMo SFT, where $\mathrm{SR}(\dW) = 382$, transfers only 1.3\% of the update's rank. The update's spectral energy is absorbed into the existing modes of $\W_0$, not stacked on top. Second, cumulative RL mirrors SFT. For example, C1 base$\to$GRPO-A ($\Delta\mathrm{SR} = -0.60$) tracks C1 base$\to$SFT ($-0.60$), so the incremental RL step adds almost no extra rank change. The same pattern holds across the larger chains: base$\to$GPPO in C2, base$\to$RL in C3, and base$\to$RL in C12 all stay close to their corresponding base$\to$SFT rank changes.

\begin{table}[ht]
  \centering
  \caption{Stable rank and additivity for all 42 transitions.
  $\mathrm{SR}_0$/$\mathrm{SR}_\Delta$/$\mathrm{SR}_1$: stable rank of
  $\W_0$/$\dW$/$\W_1$; $\alpha$: additivity ratio.}
  \label{tab:all-transitions}
  \small
  \setlength{\tabcolsep}{4pt}
  \begin{tabular}{lccccc}
  \toprule
  Transition & $\mathrm{SR}_0$ & $\mathrm{SR}_\Delta$ & $\mathrm{SR}_1$ & $\Delta$\% & $\alpha$ \\
  \midrule
  \multicolumn{6}{l}{\textit{SFT (base$\to$SFT)}} \\
  C1 & 109.8 & 82.3 & 109.2 & $-0.60$ & $-0.007$ \\
  C2 & 167.7 & 277.5 & 165.2 & $-1.49$ & $+0.017$ \\
  C3 & 183.1 & 197.2 & 183.5 & $+0.17$ & $-0.002$ \\
  C5 & 181.0 & 19.7 & 181.2 & $+0.07$ & $+0.001$ \\
  C6 & 176.6 & 10.7 & 176.3 & $-0.15$ & $-0.001$ \\
  C7 & 140.7 & 156.4 & 140.6 & $-0.09$ & $-0.001$ \\
  C8 & 154.4 & 63.7 & 154.5 & $+0.04$ & $+0.000$ \\
  C9 & 178.7 & 64.4 & 178.8 & $+0.05$ & $-0.000$ \\
  C10 & 176.6 & 59.5 & 176.7 & $+0.07$ & $+0.000$ \\
  C11 & 178.5 & 382.1 & 184.3 & $+3.21$ & $+0.013$ \\
  C12 & 133.9 & 200.4 & 130.7 & $-2.36$ & $-0.007$ \\
  \midrule
  \multicolumn{6}{l}{\textit{Incremental RL (SFT$\to$RL)}} \\
  C1: GRPO-A & 109.2 & 11.3 & 109.2 & $+0.00$ & $-0.000$ \\
  C1: GRPO-B & 109.2 & 18.6 & 109.2 & $+0.02$ & $+0.000$ \\
  C2: GPPO & 165.2 & 89.4 & 165.2 & $+0.00$ & $-0.000$ \\
  C5: DPO & 181.2 & 27.4 & 181.2 & $+0.02$ & $-0.000$ \\
  C6: DPO & 176.3 & 24.0 & 176.3 & $-0.00$ & $-0.000$ \\
  C7: DPO & 140.6 & 20.4 & 140.5 & $-0.05$ & $+0.000$ \\
  C7: Inst. & 140.6 & 21.7 & 140.5 & $-0.01$ & $+0.000$ \\
  C8: PPO & 154.5 & 19.6 & 154.4 & $-0.03$ & $+0.000$ \\
  C9: RL & 178.8 & 18.3 & 178.8 & $+0.00$ & $-0.000$ \\
  C10: PRIME & 176.7 & 10.8 & 176.7 & $+0.00$ & $+0.000$ \\
  C11: RL & 184.3 & 69.3 & 184.2 & $-0.03$ & $-0.000$ \\
  C12: RL & 130.7 & 25.7 & 130.7 & $-0.02$ & $+0.000$ \\
  \midrule
  \multicolumn{6}{l}{\textit{Cumulative RL (base$\to$RL)}} \\
  C1: GRPO-A & 109.8 & 82.3 & 109.2 & $-0.60$ & $-0.007$ \\
  C1: GRPO-B & 109.8 & 82.3 & 109.2 & $-0.58$ & $-0.007$ \\
  C2: GPPO & 167.7 & 277.5 & 165.2 & $-1.48$ & $+0.017$ \\
  C3: RL & 183.1 & 202.1 & 183.5 & $+0.18$ & $-0.002$ \\
  C4: direct GRPO & 167.7 & 192.6 & 167.6 & $-0.07$ & $+0.002$ \\
  C5: DPO & 181.0 & 20.0 & 181.2 & $+0.09$ & $+0.001$ \\
  C6: DPO & 176.6 & 10.7 & 176.3 & $-0.15$ & $-0.001$ \\
  C6: PPO & 176.6 & 29.8 & 176.6 & $+0.00$ & $-0.000$ \\
  C6: RAFT & 176.6 & 20.7 & 176.6 & $-0.01$ & $+0.000$ \\
  C7: DPO & 140.7 & 156.9 & 140.5 & $-0.13$ & $-0.001$ \\
  C7: Inst. & 140.7 & 157.0 & 140.5 & $-0.10$ & $-0.001$ \\
  C8: PPO & 154.4 & 63.5 & 154.4 & $+0.00$ & $+0.000$ \\
  C9: RL & 178.7 & 64.5 & 178.8 & $+0.05$ & $-0.000$ \\
  C10: PRIME & 176.6 & 59.5 & 176.7 & $+0.08$ & $+0.000$ \\
  C10: PRIME-Zero & 176.6 & 6.6 & 176.6 & $+0.01$ & $+0.000$ \\
  C11: RL & 178.5 & 382.1 & 184.2 & $+3.18$ & $+0.013$ \\
  C11: RL-Zero & 178.5 & 28.1 & 178.5 & $-0.02$ & $-0.000$ \\
  C12: RL & 133.9 & 200.4 & 130.7 & $-2.38$ & $-0.007$ \\
  \bottomrule
  \end{tabular}
  \end{table}

\paragraph{A note on numerical rank.}
At machine epsilon ($\sim\!1.2 \times 10^{-7}$), apparent rank changes
of up to $+1{,}439$ occur (C7 OLMo-2 \texttt{k\_proj}, layers~8--9)
because $\sim$1{,}400 very small singular values cross the threshold
after training.  At any $\varepsilon \geq 0.001$, the change drops to
$\leq 87$.  Numerical rank at machine epsilon is not useful for this
analysis.

\subsection{Interpretation}

The diagonal and off-diagonal components act within $\mathcal{S}$ and cannot change effective rank by construction. The null component accesses directions outside $\mathcal{S}$, but the rank invariance result shows that this does not create new dominant modes. Out-of-frame components introduced by $\dW_{\mathrm{null}}$ are absorbed into the spectral structure of $\W_0$ rather than expanding it. This is still a local statement about one layer. Residual paths mean a direction outside one layer's pretrained frame may already be represented elsewhere in the network. The narrower point is that $\dW_{\mathrm{null}}$ can access directions outside a given layer's pretrained frame without increasing that layer's effective dimensionality.

The broader lesson is simple. post-training can change \emph{which} directions a layer reads from and writes to without increasing \emph{how many} dominant modes it maintains. Together with the main text's finding that removing the diagonal usually does not hurt, this suggests that post-training is mainly a routing problem inside a spectral envelope fixed by pretraining.



\section{Causal Decomposition: Full Evaluation Results}
\label{app:causal-eval}

This appendix gives the full causal decomposition results summarized in Section~\ref{sec:causal_decomp}. For each transition, we reconstruct $\W_1 = \W_0 + \dW$ using either a single component or the OffNull model that keeps the off and null pair from the P-matrix decomposition, then evaluate average accuracy across five math benchmarks: MATH500, AIME24, AMC23, Minerva Math, and OlympiadBench. ``Retention'' is accuracy relative to the full model.

\paragraph{Reading the tables.}
Each row is one post-training transition. \textbf{Full} is the unmodified $\W_1$. \textbf{Diag} keeps only $\dW_{\mathrm{diag}}$. \textbf{Off} keeps only the off-diagonal component. \textbf{Null} keeps only the null space component. \textbf{OffNull} removes the diagonal and keeps both routing terms. Values near 100\% in the OffNull column mean the diagonal is not needed for downstream performance. Tables~\ref{tab:causal-sft} and~\ref{tab:causal-rl} separate SFT and RL transitions for readability.

\begin{table}[ht]
\centering
\caption{Causal decomposition: SFT transitions (base\,$\to$\,SFT).
  Accuracy (\%) averaged over five math benchmarks; retention relative
  to Full in parentheses.}
\label{tab:causal-sft}
\small
\setlength{\tabcolsep}{4pt}
\resizebox{\textwidth}{!}{%
\begin{tabular}{llcccccc}
\toprule
Chain & Base $\to$ SFT model & Base & Full & Diag & Off & Null & OffNull \\
\midrule
C1  & Qwen2.5-1.5B $\to$ DS-R1-Distill
    & 31.3 & 49.2 & 6.4\tss{13} & 31.3\tss{64} & 18.5\tss{37} & 28.9\tss{59} \\
C2  & Qwen3-8B $\to$ KlearReasoner-SFT
    & 21.2 & 60.4 & 14.0\tss{23} & 9.5\tss{16} & 9.7\tss{16} & 56.1\tss{93} \\
C3  & Qwen3-14B $\to$ UniReason-SFT
    & 17.2 & 49.1 & 28.0\tss{57} & 54.3\tss{111} & 55.7\tss{113} & 45.8\tss{93} \\
C5  & Mistral-7B $\to$ mistral-sft-beta
    & 3.6 & 5.7 & 1.9\tss{33} & 5.8\tss{102} & 6.2\tss{110} & 5.7\tss{100} \\
C6  & Qwen2.5-Math-7B $\to$ RLHFlow-SFT
    & 35.8 & 43.5 & 38.8\tss{89} & 49.7\tss{114} & 48.0\tss{110} & 42.1\tss{97} \\
C7  & OLMo-2-7B $\to$ SFT
    & 5.0 & 8.6 & 7.9\tss{92} & 9.0\tss{105} & 9.8\tss{114} & 7.5\tss{88} \\
C8 & LLaMA-3-8B $\to$ OpenRLHF-SFT
    & 4.5 & 10.0 & 2.2\tss{22} & 9.7\tss{97} & 7.6\tss{75} & 10.9\tss{109} \\
C9 & DSMath-7B $\to$ Instruct
    & 4.6 & 22.1 & 5.6\tss{25} & 19.6\tss{89} & 17.8\tss{81} & 21.0\tss{95} \\
C10 & Qwen2.5-Math-7B $\to$ Eurus2-SFT
    & 35.8 & 35.0 & 41.9\tss{120} & 39.9\tss{114} & 38.8\tss{111} & 34.0\tss{97} \\
C11 & MiMo-7B $\to$ MiMo-SFT
    & 9.6 & 65.6 & 0.0\tss{0} & 57.6\tss{88} & 44.7\tss{68} & 66.8\tss{102} \\
C12 & OLMo-3-7B $\to$ Think-SFT
    & 7.3 & 61.1 & 2.8\tss{5} & 40.9\tss{67} & 19.3\tss{32} & 62.0\tss{102} \\
\bottomrule
\end{tabular}
}

\medskip
{\footnotesize \tss{$n$} denotes retention \%.
OffNull $\geq$ 88\% for 10/11 chains: removing the diagonal preserves
most or all performance.  C1 (59\%) is the sole exception, where the
SFT is heavily distilled from DeepSeek-R1.}
\end{table}

\begin{table}[ht]
\centering
\caption{Causal decomposition: RL transitions (SFT\,$\to$\,RL or
  base\,$\to$\,RL).  Same format as Table~\ref{tab:causal-sft}.}
\label{tab:causal-rl}
\small
\setlength{\tabcolsep}{4pt}
\resizebox{\textwidth}{!}{%
\begin{tabular}{lllccccc c}
\toprule
Chain & RL model & Alg. & Base & Full & Diag & Off & Null & OffNull \\
\midrule
C1  & DeepScaleR        & GRPO & 49.2 & 51.8 & 41.0\tss{79} & 43.8\tss{85} & 51.7\tss{100} & 54.4\tss{105} \\
C1  & Nemotron          & GRPO & 49.2 & 61.3 & 40.3\tss{66} & 57.6\tss{94} & 59.2\tss{97} & 62.7\tss{102} \\
C2  & KlearReasoner     & GPPO & 60.4 & 71.7 & 58.6\tss{82} & 69.8\tss{97} & 69.0\tss{96} & 72.6\tss{101} \\
C3  & UniReason         & GRPO & 49.1 & 66.5 & 75.1\tss{113} & 64.2\tss{96} & 65.9\tss{99} & 65.0\tss{98} \\
C5  & zephyr-beta       & DPO  & 5.7 & 3.1 & 5.2\tss{168} & 5.3\tss{171} & 4.7\tss{153} & 3.7\tss{120} \\
C6  & DPO               & DPO  & 43.5 & 49.7 & 41.2\tss{83} & 45.0\tss{90} & 48.5\tss{98} & 49.8\tss{100} \\
C6  & PPO-Zero          & PPO  & 35.8 & 50.1 & 37.4\tss{75} & 43.6\tss{87} & 44.4\tss{89} & 49.7\tss{99} \\
C6  & RAFT-Zero         & RAFT & 35.8 & 44.6 & 38.4\tss{86} & 43.1\tss{97} & 44.3\tss{99} & 45.4\tss{102} \\
C7  & DPO               & DPO  & 8.6 & 11.3 & 8.8\tss{79} & 9.3\tss{83} & 10.5\tss{94} & 11.7\tss{104} \\
C7  & RLVR              & RLVR & 11.3 & 11.7 & 11.9\tss{101} & 10.4\tss{89} & 12.2\tss{104} & 10.8\tss{93} \\
C8 & PPO               & PPO  & 10.0 & 11.1 & 10.8\tss{97} & 10.1\tss{91} & 12.1\tss{109} & 11.2\tss{101} \\
C9 & RL                & GRPO & 22.1 & 25.5 & 19.2\tss{75} & 22.4\tss{88} & 21.2\tss{83} & 23.8\tss{93} \\
C10 & PRIME             & PRIME & 35.0 & 44.2 & 36.1\tss{82} & 39.7\tss{90} & 40.7\tss{92} & 44.3\tss{100} \\
C10 & PRIME-Zero        & PRIME & 35.8 & 42.1 & 35.0\tss{83} & 36.4\tss{86} & 41.2\tss{98} & 38.6\tss{92} \\
C11 & MiMo-RL           & RL   & 65.6 & 69.1 & 67.3\tss{97} & 70.2\tss{102} & 69.7\tss{101} & 67.0\tss{97} \\
C11 & MiMo-RL-Zero      & RL   & 9.6 & 49.7 & 10.5\tss{21} & 36.8\tss{74} & 39.9\tss{80} & 47.3\tss{95} \\
C12 & Think-RL          & RL   & 61.1 & 68.2 & 61.2\tss{90} & 65.5\tss{96} & 64.9\tss{95} & 67.8\tss{99} \\
\bottomrule
\end{tabular}
}

\medskip
{\footnotesize OffNull $\geq$ 92\% for 15/17 RL transitions.
C5 zephyr-beta (120\%) and C7 RLVR (93\%) are the only cases
below 95\%; both involve very small relative updates where all components
are near noise level.} 
\end{table}

\paragraph{Summary statistics.}
Across all 28 transitions, OffNull retention has median 99\% and mean 97\%. On the 25 positive gain transitions used for the main text summaries, the OffNull model stays within 4.3 points of the full model in 24 cases. Across the full set, it is at least 95\% in 22 of 28 transitions and at least 92\% in 25 of 28. Diag only retention is much lower, with median 79\% for RL and 23\% for SFT. In 6 of 11 SFT chains, the diagonal alone retains at most 25\% of the full performance. Null only retention is generally higher than diagonal only retention for both SFT and RL, which matches the routing interpretation: null space changes contribute more to behavioral change than simple spectral rescaling.



\section{Energy Fraction Decomposition: Full Results}
\label{app:energy-fractions}

The main text shows that the diagonal component usually carries only a small fraction of the update energy, while the off-diagonal and null space terms carry most of it. This appendix gives the full transition by transition breakdown and explains where that split comes from.

\paragraph{Definition.}
For each transition we compute the energy fraction carried by each
component:
$\fdiag = \|\dW_{\mathrm{diag}}\|_F^2 / \|\dW\|_F^2$, and
analogously $f_{\mathrm{off}}$ and $f_{\mathrm{null}}$.  By the
Pythagorean property of the decomposition, the three fractions sum
to~1.

\paragraph{Geometric constraint on the partition.}
For square projections ($m = n = r$), both $\col(\U_0)$ and $\col(\V_0)$ span their ambient spaces, so $\mathcal{S} = \R^{m \times n}$ and $\dW_{\mathrm{null}}$ vanishes identically. In that case the decomposition reduces to the in frame part studied by prior spectral analyses of post-training~\citep{zhu2025pathtakenrlvrprovably, he2026understandingposttrainingstructuralchanges}. In practice, however, five of the seven projections per transformer layer are rectangular (gate, up, down, and $k$/$v$ under GQA), and the null component typically carries 26--56\% of the update energy.


\paragraph{Full results.}
Table~\ref{tab:energy-fractions} reports the mean energy fraction
across all layers and module types for every transition.

\begin{table}[ht]
\centering
\caption{Mean energy fraction of each $\dW$ component, averaged over
  all layers and module types.  Fractions sum to~1 by construction.}
\label{tab:energy-fractions}
\small
\setlength{\tabcolsep}{5pt}
\begin{tabular}{llccc}
\toprule
Chain & Transition & $f_{\mathrm{diag}}$ & $f_{\mathrm{off}}$ & $f_{\mathrm{null}}$ \\
\midrule
\multicolumn{5}{l}{\textit{SFT (base$\to$SFT)}} \\
C1  & Qwen2.5-1.5B $\to$ DS-R1-Distill  & 0.204 & 0.288 & 0.508 \\
C2  & Qwen3-8B $\to$ KlearReasoner-SFT   & 0.149 & 0.456 & 0.397 \\
C3  & Qwen3-14B $\to$ UniReason-SFT      & 0.011 & 0.504 & 0.487 \\
C5  & Mistral-7B $\to$ mistral-sft-beta  & 0.000 & 0.508 & 0.492 \\
C6  & Qwen2.5-Math-7B $\to$ RLHFlow-SFT  & 0.000 & 0.453 & 0.548 \\
C7  & OLMo-2-7B $\to$ SFT                & 0.000 & 0.736 & 0.265 \\
C8 & LLaMA-3-8B $\to$ OpenRLHF-SFT      & 0.000 & 0.499 & 0.501 \\
C9 & DSMath-7B $\to$ Instruct            & 0.007 & 0.734 & 0.260 \\
C10 & Qwen2.5-Math-7B $\to$ Eurus2-SFT   & 0.000 & 0.444 & 0.556 \\
C11 & MiMo-7B $\to$ MiMo-SFT             & 0.005 & 0.525 & 0.471 \\
C12 & OLMo-3-7B $\to$ Think-SFT          & 0.003 & 0.735 & 0.263 \\
\midrule
\multicolumn{5}{l}{\textit{RL (both incremental SFT$\to$RL and base$\to$RL)}} \\
C1  & $\to$ DeepScaleR (GRPO)             & 0.001 & 0.450 & 0.549 \\
C1  & $\to$ Nemotron (GRPO)               & 0.001 & 0.438 & 0.562 \\
C2  & $\to$ KlearReasoner (GPPO)           & 0.001 & 0.531 & 0.469 \\
C3  & $\to$ UniReason (GRPO)               & 0.011 & 0.504 & 0.487 \\
C4 & $\to$ SkyRL-Agent (direct GRPO) & 0.033 & 0.518 & 0.451 \\
C5  & $\to$ zephyr-beta (DPO)              & 0.000 & 0.507 & 0.494 \\
C6  & $\to$ DPO                            & 0.000 & 0.416 & 0.585 \\
C6  & $\to$ PPO-Zero                       & 0.000 & 0.427 & 0.573 \\
C6  & $\to$ RAFT-Zero                      & 0.000 & 0.441 & 0.559 \\
C7  & $\to$ DPO                            & 0.001 & 0.737 & 0.264 \\
C7  & $\to$ RLVR                           & 0.000 & 0.734 & 0.268 \\
C8 & $\to$ PPO                            & 0.000 & 0.488 & 0.513 \\
C9 & $\to$ RL (GRPO)                      & 0.000 & 0.739 & 0.263 \\
C10 & $\to$ PRIME                          & 0.000 & 0.460 & 0.541 \\
C10 & $\to$ PRIME-Zero                     & 0.000 & 0.456 & 0.545 \\
C11 & $\to$ MiMo-RL                        & 0.005 & 0.539 & 0.457 \\
C11 & $\to$ MiMo-RL-Zero                   & 0.002 & 0.533 & 0.466 \\
C12 & $\to$ Think-RL                       & 0.000 & 0.738 & 0.263 \\
\bottomrule
\end{tabular}
\end{table}

\paragraph{Observations.}
Three patterns are clear across the 28 transitions. First, the diagonal is negligible for RL. All 17 RL transitions have $\fdiag \leq 0.011$, and 14 of 17 have $\fdiag \leq 0.001$. RL places almost all of its update energy in off-diagonal rotation and null space redirection. Second, SFT shows a wider range. Two SFT chains, C1 and C2, allocate 15--20\% of their energy to the diagonal. These are the only transitions with substantial $\fdiag$, and both involve aggressive distillation with large relative updates. The other 9 SFT chains have $\fdiag \leq 0.011$, which is indistinguishable from RL. Third, the split between off-diagonal and null space is governed mostly by matrix shape, not by algorithm. Chains C7, C9, and C12 consistently show $f_{\mathrm{off}} \approx 0.73$ and $f_{\mathrm{null}} \approx 0.26$ whether the transition is SFT or RL. By contrast, chains with more rectangular projections, such as C1, C6, and C10, show $f_{\mathrm{null}} > 0.5$.


\section{Cross-Method Comparisons from a Shared Base}
\label{sec:cross_model}

The sequential analysis above compared SFT and RL inside one pipeline. Here we ask a different question: when different groups post-train the \emph{same} base model with different algorithms, data, or recipes, how similar are the resulting updates?

\begin{figure}[t]
  \centering
  \includegraphics[width=\textwidth]{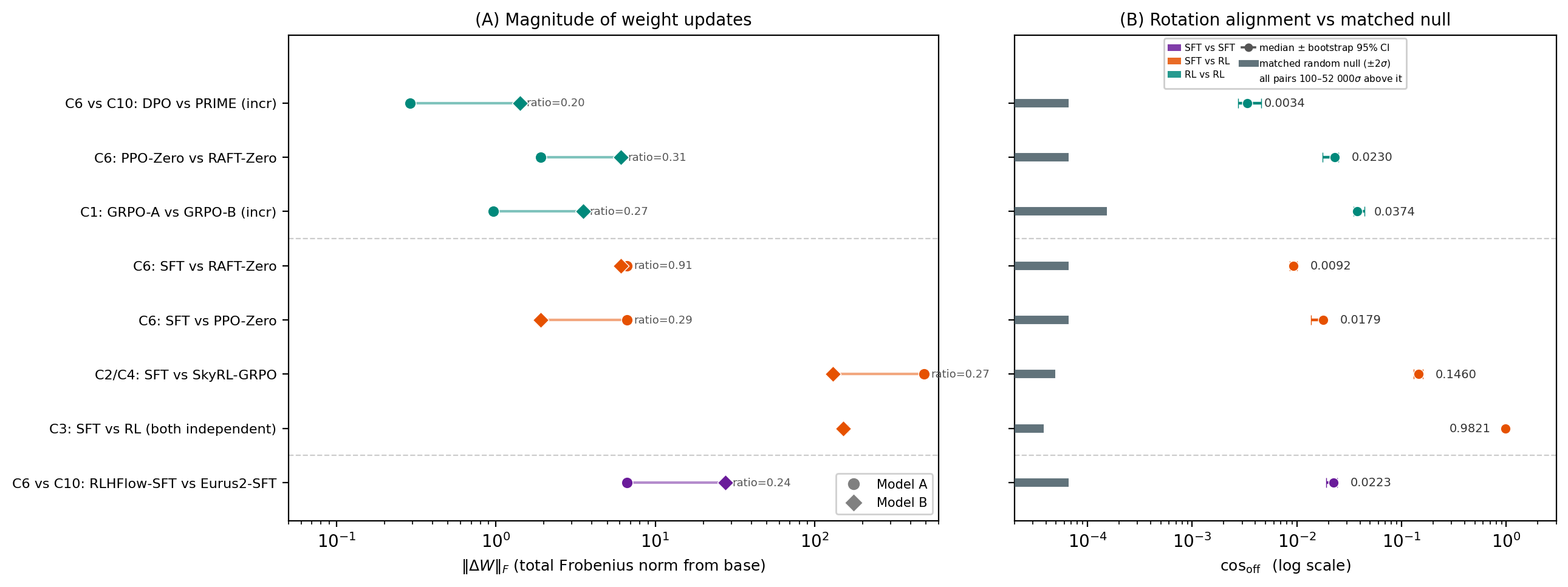}
  \caption{
    \textbf{Cross method comparison: same base, different training.}
    \textbf{(A)}~Frobenius norms of the two updates (log scale), with
    the smaller to larger ratio annotated.
    \textbf{(B)}~Off-diagonal cosine similarity. The matched isotropic null accounts for each pair's matrix shapes and the aggregation statistic used in the figure. All pairs are positively aligned beyond this null. The sole exception is C3, where independently trained SFT and RL produce nearly identical rotations ($\cos = 0.98$) and closely matched update magnitudes.}
  \label{fig:cross_model}
\end{figure}

This comparison is possible because GRRR projects all updates into the same pretrained SVD basis. We study eight pairs across three settings: RL against RL, SFT against RL, and SFT against SFT from a shared base. For each pair, we report the Frobenius norm of each update and the off-diagonal cosine between their P-matrices. Results appear in Figure~\ref{fig:cross_model}.

\paragraph{Independent methods show low but structured alignment.}
Excluding C3, the off-diagonal cosine ranges from 0.003 to 0.146. Figure~\ref{fig:cross_model}B shows the two-sigma random baseline in gray. Every pair lies above that envelope. The observed values are small in absolute terms but they are not random. This suggests that independent post-training methods share a weak but real directional bias from the base model and task domain. The pattern appears in all three comparison types: different RL algorithms from the same checkpoint, SFT compared against independent RL from the same base, and different SFT recipes from the same base.

Comparison type itself does not predict alignment very well. RL against RL pairs range from 0.003 to 0.037, SFT against RL pairs range from 0.009 to 0.146, and the single SFT against SFT pair is 0.022. The variation within each category is comparable to the variation between categories. This suggests that data and training scale matter more than the SFT/RL label when determining update direction.

\paragraph{C3: controlled data and compute produce identical updates.}
The striking exception is C3 (Qwen3-14B), where UniReason-SFT and UniReason-RL are trained independently from the same base with controlled data and computation. Their off-diagonal cosine is 0.982, and Figure~\ref{fig:cross_model}A shows that their update norms are also closely matched. The updates are nearly identical in both direction and magnitude even though one uses SFT and the other uses GRPO. This shows that when training data and optimization budget are matched, the algorithm label has almost no effect on update direction. The base model and data determine \emph{where} post-training moves in weight space. The algorithm mostly determines \emph{how} it gets there.

\paragraph{Comparison with the sequential pipeline.}
These cross method results complement the sequential analysis. In sequential pipelines, where RL starts from the SFT checkpoint, the off-diagonal cosine stays below 0.008. In shared base comparisons, the cosine is higher, between 0.003 and 0.146. The reason is geometric. Sequential orthogonality appears because RL is initialized from SFT and therefore moves into directions that SFT left unused. Cross method alignment is low for a different reason: independent training trajectories diverge in high dimensions even when they share some directional bias from the same base.

The C3 result ties these observations together. When data and compute are controlled, even the SFT/RL distinction largely disappears. The near orthogonality of the sequential pipeline should therefore not be read as an intrinsic algorithmic difference. It mainly reflects the fact that RL starts from the SFT checkpoint and is constrained to a different part of the loss landscape.

\section{Random subspace study}
\label{app:rando}

\begin{table}[h]
\centering
\small
\begin{tabular}{lrrrrr}
\toprule
Transition
& Full
& Remove diag.
& Remove random
& $\Delta_{\mathrm{diag}}$
& $\Delta_{\mathrm{rand}}$ \\
\midrule
C1 SFT & 49.2 & 28.9 & 44.2 & $-20.3$ & $-5.0$ \\
C2 SFT & 60.4 & 56.1 & 45.2 & $-4.3$ & $-15.2$ \\
\bottomrule
\end{tabular}
\caption{Removing the learned diagonal component compared with removing a random component of the same Frobenius norm within the pretrained SVD core. Random-removal results are averaged over three seeds.}
\label{tab-remove-diag}
\end{table}

\end{document}